%% file: iclr2025_conference.tex
\documentclass{article} 
\usepackage{iclr2025_conference,times}

\input{math_commands.tex}

\usepackage{hyperref}
\usepackage{url}

\usepackage{amsmath}
\usepackage{amssymb}
\usepackage{graphicx}
\usepackage{multirow}
\usepackage{booktabs}
\usepackage{subcaption}
\usepackage{rotating}

\usepackage{marvosym} 

\makeatletter
\renewcommand\@fnsymbol[1]{%
  \ifcase#1\relax
    \or *
    \or \mbox{\scriptsize\Letter} 
    \or 1
    \or 2
    \or 3
    \or 4
  \fi
}
\makeatother

\title{Rethinking the generalization of drug target affinity prediction algorithms via similarity aware evaluation}

\author{Chenbin Zhang\thanks{Equal contribution.}\hspace{0.35em}\footnotemark[3]\hspace{0.4em}, Zhiqiang Hu\footnotemark[1]\hspace{0.4em}\thanks{Corresponding Author.}\hspace{0.65em}\footnotemark[4]\hspace{0.4em}, Chuchu Jiang\footnotemark[1]\hspace{0.4em}\footnotemark[5]\hspace{0.4em}, Wen Chen\footnotemark[6]\hspace{0.4em}, Jie Xu\footnotemark[5]\hspace{0.4em}, Shaoting Zhang\footnotemark[5] \\
\\
\footnotemark[3]\hspace{0.4em}MoleculeMind, \footnotemark[4]\hspace{0.4em}Peking University, \footnotemark[5]\hspace{0.4em}Shanghai AI Laboratory, \footnotemark[6]\hspace{0.4em}SenseTime Research\\
\\
\texttt{chenbinzhang@moleculemind.com,huzq@pku.edu.cn,} \\
\texttt{\{jiangchuchu,xujie,zhangshaoting\}@pjlab.org.cn,} \\
\texttt{chenwen@sensetime.com}
}

\iclrfinalcopy 
\begin{document}

\maketitle

\begin{abstract}
Drug-target binding affinity prediction is a fundamental task for drug discovery. It has been extensively explored in literature and promising results are reported. However, in this paper, we demonstrate that the results may be misleading and cannot be well generalized to real practice. The core observation is that the canonical randomized split of a test set in conventional evaluation leaves the test set dominated by samples with high similarity to the training set. The performance of models is severely degraded on samples with lower similarity to the training set but the drawback is highly overlooked in current evaluation. As a result, the performance can hardly be trusted when the model meets low-similarity samples in real practice. To address this problem, we propose a framework of similarity aware evaluation in which a novel split methodology is proposed to adapt to any desired distribution. This is achieved by a formulation of optimization problems which are approximately and efficiently solved by gradient descent. We perform extensive experiments across five representative methods in four datasets for two typical target evaluations and compare them with various counterpart methods. Results demonstrate that the proposed split methodology can significantly better fit desired distributions and guide the development of models. Code is released at \url{https://github.com/Amshoreline/SAE/tree/main}.
\end{abstract}

\section{Introduction}
Drug-target binding affinity (DTA) prediction is a fundamental and crucial task for drug discovery. It evaluates the effectiveness of drug candidates, or samples, and sees its application in a large-scale virtual screening where most ineffective candidates are filtered out to save experimental cost and time \citep{chatterjee2023improving}. DTA is quantitatively measured by inhibition constant Ki, half maximal inhibitory concentration IC50, etc., which are all real-valued \citep{monteiro2022explainable}. The prediction performance is commonly evaluated by mean absolute error (MAE) and coefficient of determination (R$^2$).

The task of DTA prediction has been extensively studied for decades \citep{chen2018rise,askr2023deep}. Related works can be categorized as structure-based, sequence-based, and similarity-based \citep{wu2018moleculenet,chuang2020learning}. Structure-based methods rely on 3D structures of samples, target proteins, or their complexes. Although theoretically accessible to most comprehensive information following the dogma ``structure determines function'', structure-based methods are limited by available 3D structures, especially experimentally verified structures, and also hindered in practice by poor time efficiency. In contrast, sequence-based and similarity-based methods are fast and do not set 3D structures as prerequisite \citep{xu2017seq2seq,zhang2022pushing}. Instead, they take as input residual sequences, Simplified Molecular-Input Line-Entry System (SMILES) sequences, fingerprint sequences, atom-bond graphs, or the derived pairwise similarities, which are easier to acquire with lower cost.
Moreover, these sequences and similarities are readily processed by diversified sophisticated backbones including convolutional neural networks (CNNs) \citep{ozturk2018deepdta,li2019deepatom,hu2023sam}, recurrent neural networks (RNNs) \citep{karimi2019deepaffinity,yuan2022fusiondta}, graph neural networks (GNNs) \citep{nguyen2021graphdta,yang2022mgraphdta,tang2022fmgnn,wang2022molecular} and transformers \citep{chithrananda2020ChemBERTa,zhao2022attentiondta,song2023double,jiang2023pharmacophoric}, and enjoy the benefits of the development of deep learning techniques. As a result, sequence-based and similarity-based methods are shown to reach new high performance and are drawing increasing attention.

\begin{figure}[t]
  \centering
  \includegraphics[width=0.95\textwidth]{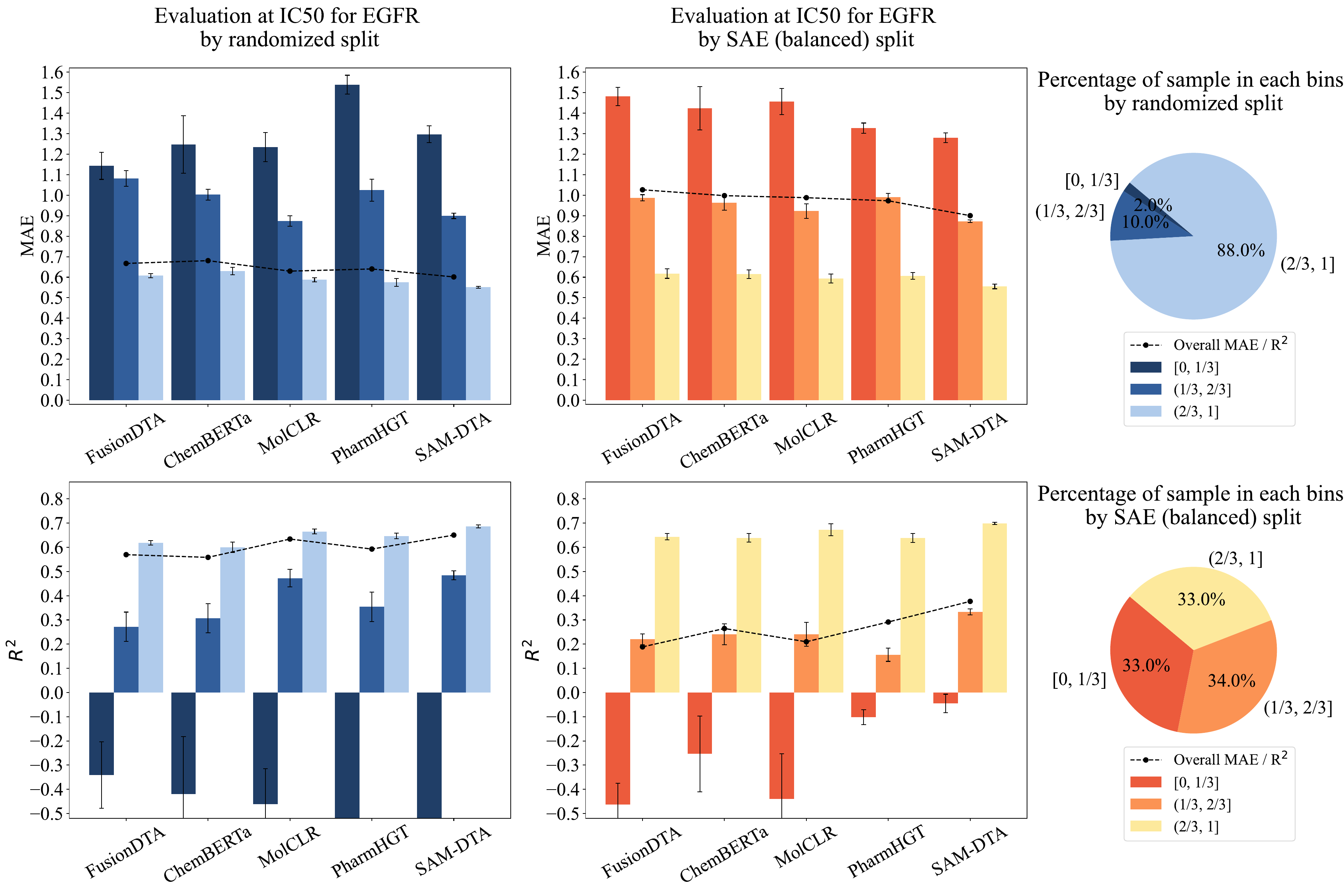}
  \caption{Comparison of randomized split and SAE (balanced) split at IC50 for EGFR. The randomized split led to 88\% of test samples yielding a high similarity ($> 2/3$) to the training set. In contrast, our SAE (balanced) split strategy ensures a more balanced distribution of similarities. The evaluation of five DTA prediction methods demonstrates that the performance aligns with the similarity levels. In the randomized split, the overall performance closely resembles that of high-similarity samples, thus failing to evaluate the performance when encountering low-similarity samples.}
  \label{fig:benchmark_EGFR}
\end{figure}

Although promising results are reported, we find, surprisingly, that these results may be misleading. Take the task of IC50 prediction for target EGFR as an example, as shown in Figure~\ref{fig:benchmark_EGFR}, we evaluate five state-of-the-art and representative methods and the best-performing one, SAM-DTA~\citep{hu2023sam}, achieves a MAE of 0.6012 and R$^2$ of 0.6505 for the \emph{whole} test set. However, if we dive into the performance and divide the test set according to the similarity of the sample to the training set, we find a clear performance degradation for low-similarity samples: the MAE deteriorates to 1.2970 for samples with similarity less than 1/3 and R$^2$ to -0.6385. The gap is huge. Nevertheless, poor performance on low-similarity samples does not affect the \emph{whole} performance since they only occupy a negligible proportion: only 16 samples with similarity less than 1/3 out of a total of 873 samples in the test set (Figure~\ref{fig:benchmark_EGFR}).
In other words, the test set is dominated by high-similarity samples and performance for low-similarity samples is overwhelmed in current evaluations. We will show that the phenomenon exists across different similarity measures, performance metrics, datasets, and methods, and therefore it is general. Consequently, the evaluation will be misleading to practitioners, especially when the trained model meets low-similarity samples when used in real practice.

We argue that the core of the problem lies in the canonical randomized split of the test set. Randomized split follows the assumption of independent identical distribution (I.I.D.), which is the foundation of most statistical learning theories. However, in drug discovery samples are not necessarily independent of each other: in practice, mutually similar variants are more likely to be tested together in high-throughput experiments, while at the same time, they have to avoid high similarity to approved drugs for intellectual property issues \citep{harren2024modern}. Empirically, drugs developed at different times show significant distinction in their properties \citep{sheridan2022prediction}.
As a result, practitioners would not always expect the samples they are going to test to follow the same distribution as historical samples. This in turn raises a request to model development that the test set should satisfy a desired distribution. For example, one may need a test set that is uniform at different similarity bins; others may ask the test set samples to be all limited within predefined similarity bounds, and so on \citep{li2017structural,simm2021splitting,luo2024enhancing,tricarico2024construction}.

We formulate the test set split with a desired distribution as a combination optimization problem.
The problem is infeasible to solve for optimum due to efficiency issues. In this work, we address this challenge by relaxing it to a continuous optimization problem where samples are allowed to coexist in the training set and test set with a ``probability'' or weight. Further, the objective function contains non-differentiable operations including taking the maximum and counting in similarity bins, and are approximated in this work by differentiable counterparts. We will show that the degree of approximation is adjustable by introduced hyper-parameters. Next, the resulting optimization problem has no closed-form solution, and hence we have resorted to Lagrangian multipliers with a numerical method implemented by PyTorch and Cooper~\citep{gallegoPosada2022cooper}.
Finally, we analyze the continuously-valued solution and find the non-negligible approximation error induced by the relaxation. To this end, we introduce a regularization term that penalizes samples whose weight is far from bipartition. We refer to our strategy as \textbf{S}imilarity \textbf{A}ware \textbf{E}valuation, abbreviated as SAE. By doing all this, we have managed to achieve test set splits with various desired distributions.

Extensive experiments are performed to substantiate the effectiveness of our split strategy.
To begin with, our split strategy can achieve a uniformly distributed test set across various similarity bins (Figure~\ref{fig:benchmark_EGFR}). Subsequently, we evaluate the performance of five DTA prediction methods on this test set. The results underscore a distinct relationship between the performance and the corresponding similarity levels, suggesting a more comprehensive assessment of balanced split across varied methods in comparison to the strategy of randomized split.
Moreover, in scenarios where the samples practitioners intend to test deviate from the distribution of existing samples, our split strategy can effectively split the training set and internal test set based on the similarity distribution of the test set (Figure~\ref{fig:mimic_split_bar}). Here we conduct hyper-parameter searches on the internal test set across five DTA prediction methods to assess the efficacy of our split strategy. Compared to previous split strategies, our split strategy facilitates the selection of optimal hyper-parameters, enhancing performance on the external test set (Figure~\ref{fig:mimic_split_score}).
In other scenarios where practitioners specify predefined similarity constraints for the test set samples, such as a maximum similarity limit of 0.4 or 0.6, or even a range bounded by a minimum and maximum similarity of 0.4 and 0.6, our split strategy ensures the majority of samples in the test set adhering to these requirements (Figure~\ref{fig:other_app_bar}).

\begin{table}[t]
\centering
\caption{Variations of $\mathit{SimilarityToTrainingSet}$ related to feature extraction, similarity measure, aggregation functions, and performance metrics. We choose PharmHGT and SAM-DTA as the example methods for the detailed showcase.}
\label{tab:random_split}
\begin{tabular}{cc@{\hspace{2mm}}c@{\hspace{2mm}}cc@{\hspace{2mm}}c@{\hspace{2mm}}c}
\toprule

\multicolumn{7}{c}{Randomized Split (MAE)} \\
\midrule
\multirow{2}{*}{Bin} & \multicolumn{3}{c}{Feature: RDKit fingerprint} & \multicolumn{3}{c}{Feature:  Avalon fingerprint} \\
\cmidrule(r){2-4} \cmidrule(r){5-7}
& Count (Ratio) & PharmHGT & SAM-DTA & Count (Ratio) & PharmHGT & SAM-DTA \\
\midrule
{[}0\phantom{/3}, 1/3] & \phantom{00}8 (0.0092) & 1.7551 & 1.6787 & \phantom{00}0 (0.0000) & - & - \\
{(}1/3, 2/3] & \phantom{0}34 (0.0389) & 1.3214 & 1.0040 & \phantom{0}28 (0.0321) & 1.4646 & 1.3319 \\
{(}2/3, \phantom{1/}1] & 831 (0.9519) & 0.6015 & 0.5743 & 845 (0.9679) & 0.6128 & 0.5770 \\
overall & 873 (1.0000) & 0.6401 & 0.6012 & 873 (1.0000) & 0.6401 & 0.6012 \\
\midrule
 & \multicolumn{3}{c}{SimilarityMeasure: Sokal similarity} & \multicolumn{3}{c}{SimilarityMeasure: Dice coefficient} \\
\midrule
{[}0\phantom{/3}, 1/3] & \phantom{0}33 (0.0378) & 1.5051 & 1.2444 & \phantom{00}0 (0.0000) & - & - \\
{(}1/3, 2/3] & 398 (0.4559) & 0.7066 & 0.6619 & \phantom{0}33 (0.0378) & 1.5051 & 1.2444 \\
{(}2/3, \phantom{1/}1] & 442 (0.5063) & 0.5157 & 0.4985 & 840 (0.9622) & 0.6061 & 0.5759 \\
overall & 873 (1.0000) & 0.6401 & 0.6012 & 873 (1.0000) & 0.6401 & 0.6012 \\
\midrule
 & \multicolumn{3}{c}{Aggregation: Top-3} & \multicolumn{3}{c}{Aggregation: Top-5} \\
\midrule
{[}0\phantom{/3}, 1/3] & \phantom{0}17 (0.0195) & 1.5188 & 1.3149 & \phantom{0}24 (0.0275) & 1.5627 & 1.3469 \\
{(}1/3, 2/3] & 171 (0.1959) & 0.8890 & 0.7748 & 240 (0.2749) & 0.8014 & 0.7228 \\
{(}2/3, \phantom{1/}1] & 685 (0.7847) & 0.5562 & 0.5401 & 609 (0.6976) & 0.5402 & 0.5239 \\
overall & 873 (1.0000) & 0.6401 & 0.6012 & 873 (1.0000) & 0.6401 & 0.6012 \\
\midrule
\multicolumn{7}{c}{Randomized Split (R$^2$)} \\
\midrule
\multirow{2}{*}{Bin} & \multicolumn{3}{c}{Feature: RDKit fingerprint} & \multicolumn{3}{c}{Feature:  Avalon fingerprint} \\
\cmidrule(r){2-4} \cmidrule(r){5-7}
& Count (Ratio) & PharmHGT & SAM-DTA & Count (Ratio) & PharmHGT & SAM-DTA \\
\midrule
{[}0\phantom{/3}, 1/3] & \phantom{00}8 (0.0092) & 0.1555 & 0.2579 & \phantom{00}0 (0.0000) & - & - \\
{(}1/3, 2/3] & \phantom{0}34 (0.0389) & -0.0371 & 0.3529 & \phantom{0}28 (0.0321) & 0.1028 & 0.2921 \\
{(}2/3, \phantom{1/}1] & 831 (0.9519) & 0.6327 & 0.6706 & 845 (0.9679) & 0.6169 & 0.6672 \\
overall & 873 (1.0000) & 0.5928 & 0.6505 & 873 (1.0000) & 0.5928 & 0.6505 \\
\midrule
 & \multicolumn{3}{c}{SimilarityMeasure: Sokal similarity} & \multicolumn{3}{c}{SimilarityMeasure: Dice coefficient} \\
\midrule
{[}0\phantom{/3}, 1/3] & \phantom{0}33 (0.0378) & -0.1562 & 0.1866 & \phantom{00}0 (0.0000) & - & - \\
{(}1/3, 2/3] & 398 (0.4559) & 0.5412 & 0.6057 & \phantom{0}33 (0.0378) & -0.1562 & 0.1866 \\
{(}2/3, \phantom{1/}1] & 442 (0.5063) & 0.6942 & 0.7156 & 840 (0.9622) & 0.6290 & 0.6711 \\
overall & 873 (1.0000) & 0.5928 & 0.6505 & 873 (1.0000) & 0.5928 & 0.6505 \\
\midrule
 & \multicolumn{3}{c}{Aggregation: Top-3} & \multicolumn{3}{c}{Aggregation: Top-5} \\
\midrule
{[}0\phantom{/3}, 1/3] & \phantom{0}17 (0.0195) & -0.6280 & -0.3148 & \phantom{0}24 (0.0275) & -0.2379 & 0.0891 \\
{(}1/3, 2/3] & 171 (0.1959) & 0.4283 & 0.5743 & 240 (0.2749) & 0.4756 & 0.5794 \\
{(}2/3, \phantom{1/}1] & 685 (0.7847) & 0.6483 & 0.6712 & 609 (0.6976) & 0.6584 & 0.6804 \\
overall & 873 (1.0000) & 0.5928 & 0.6505 & 873 (1.0000) & 0.5928 & 0.6505 \\

\bottomrule 
\end{tabular}
\end{table}

\section{Problem of Randomized Split}
\label{sec:prob_random}

In this section, we will show that the imbalanced distribution of samples and the consequent overshadowing of low-similarity samples is general for a randomized split of a test set across different similarity measures, performance metrics, datasets, and methods. 
We will first give the details of the example in Figure~\ref{fig:benchmark_EGFR} and then explore possible variants.

In the example demonstrated by Figure~\ref{fig:benchmark_EGFR}, we take the IC50 dataset of target EGFR with a total of 4,361 samples, which is one of the largest datasets we are able to find. The dataset has originated from the BindingDB database, and IC50 has been converted to its negative logarithm form, $\mathit{pIC50} = -\log_{10} \mathit{ IC50\ (Molar)}$ following the convention. Then, we randomly split the dataset into a training set and test set with a ratio of 8:2. Subsequently, we train and validate the DTA prediction models on the training set and evaluate their performance on the test set. It should be noted that we follow the original hyper-parameter tuning procedure in the source code of each DTA prediction method.

To perform the fine-grained evaluation with respect to the similarity, we first define the pairwise similarity for sample $x_1$ and $x_2$,
\begin{equation}
\mathit{PairwiseSimlarity}(x_1, x_2) = \mathit{SimilarityMeasure}\left(\mathit{Feature}(x_1), \mathit{Feature}(x_2) \right)
\end{equation}
and then derive the similarity to the union of the training set by aggregation, for sample $x\in \mathit{TestSet}$.

\begin{equation}
\mathit{SimlarityToTrainingSet}(x) = \mathop{\mathit{Aggregation}}_{t\in \mathit{TrainingSet}}\ \mathit{PairwiseSimlarity}(x, t)
\end{equation}
where in the example of Figure~\ref{fig:benchmark_EGFR}, we set $\mathit{Feature}$ as the Morgan fingerprint and $\mathit{SimilarityMeasure}$ as the Tanimoto coefficient, which are both commonly used to measure the similarity of samples, and we set $\mathit{Aggregation}$ as the maximum function \citep{bajusz2015tanimoto,ying2021improving}.

Next, we compare other variants for these functions. For the feature extractor $\mathit{Feature}$, we compare other widely used molecular descriptors including Avalon fingerprint and RDKit fingerprint (a.k.a. topological fingerprint); for the function $\mathit{SimilarityMeasure}$ we compare Sokal similarity and Dice coefficient, which are both symmetric for its parameters; and finally for the $\mathit{Aggregation}$ function, we compare the general top-$k$ averaging where the maximum function can be seen as a special case of $k=1$. Note that averaging or taking the median over the whole training set is not suitable. This is because the majority of samples in the training set have a low similarity to a specific sample, and averaging or taking the median over the whole training set is not able to tell whether there exists any high-similarity ones. The results are shown in Table~\ref{tab:random_split}. Here we choose two example methods for detailed showcase (PharmHGT~\citep{jiang2023pharmacophoric} and SAM-DTA~\citep{hu2023sam}), while the results of other methods can be found in the appendix.

\begin{table}[t]
\centering
\caption{Comparison of randomized split and SAE (balanced) split at IC50 for BACE1, Ki for Carbonic anhydrase I and Carbonic anhydrase II. We choose PharmHGT and SAM-DTA as the example methods for the detailed showcase.}
\label{tab:benchmark}
\begin{tabular}{cc@{\hspace{2mm}}c@{\hspace{2mm}}cc@{\hspace{2mm}}c@{\hspace{2mm}}c}
\toprule

\multicolumn{7}{c}{IC50 for Target BACE1 (MAE)} \\
\midrule
\multirow{2}{*}{Bin} & \multicolumn{3}{c}{Randomized Split}     & \multicolumn{3}{c}{SAE (balanced) Split}  \\
\cmidrule(r){2-4} \cmidrule(r){5-7}
& Count (Ratio) & PharmHGT & SAM-DTA & Count (Ratio) & PharmHGT & SAM-DTA \\
\midrule
{[}0\phantom{/3}, 1/3] & \phantom{0}10 (0.0108) & 1.3743 & 1.1204 & 309 (0.3330) & 1.1397 & 1.0309 \\
{(}1/3, 2/3] & \phantom{0}67 (0.0722) & 0.6334 & 0.6928 & 311 (0.3351) & 0.6410 & 0.6693 \\
{(}2/3, \phantom{1/}1] & 851 (0.9170) & 0.4611 & 0.4594 & 308 (0.3319) & 0.4747 & 0.4808 \\
overall & 928 (1.0000) & 0.4834 & 0.4834 & 928 (1.0000) & 0.7518 & 0.7272 \\
\midrule
\multicolumn{7}{c}{IC50 for Target BACE1 (R$^2$)} \\
\midrule
{[}0\phantom{/3}, 1/3] & \phantom{0}10 (0.0108) & -0.0553 & 0.3702 & 309 (0.3330) & -0.2983 & -0.1261 \\
{(}1/3, 2/3] & \phantom{0}67 (0.0722) & 0.6789 & 0.6439 & 311 (0.3351) & 0.5848 & 0.5637 \\
{(}2/3, \phantom{1/}1] & 851 (0.9170) & 0.7113 & 0.7150 & 308 (0.3319) & 0.7797 & 0.7803 \\
overall & 928 (1.0000) & 0.7190 & 0.7256 & 928 (1.0000) & 0.5329 & 0.5665 \\
\midrule
\multicolumn{7}{c}{Ki for Target Carbonic anhydrase I (MAE)} \\
\midrule
\multirow{2}{*}{Bin} & \multicolumn{3}{c}{Randomized Split}     & \multicolumn{3}{c}{SAE (balanced) Split}  \\
\cmidrule(r){2-4} \cmidrule(r){5-7}
& Count (Ratio) & PharmHGT & SAM-DTA & Count (Ratio) & PharmHGT & SAM-DTA \\
\midrule
{[}0\phantom{/3}, 1/3] & \phantom{00}7 (0.0079) & 1.1467 & 0.8798 & 264 (0.2983) & 0.8410 & 0.8729 \\
{(}1/3, 2/3] & 205 (0.2316) & 0.5843 & 0.6605 & 311 (0.3514) & 0.6706 & 0.6877 \\
{(}2/3, \phantom{1/}1] & 673 (0.7605) & 0.4986 & 0.4896 & 310 (0.3503) & 0.6039 & 0.5740 \\
overall & 885 (1.0000) & 0.5236 & 0.5323 & 885 (1.0000) & 0.6981 & 0.7031 \\
\midrule
\multicolumn{7}{c}{Ki for Target Carbonic anhydrase I (R$^2$)} \\
\midrule
{[}0\phantom{/3}, 1/3] & \phantom{00}7 (0.0079) & -0.3232 & 0.1308 & 264 (0.2983) & -0.0389 & -0.0282 \\
{(}1/3, 2/3] & 205 (0.2316) & 0.5733 & 0.4820 & 311 (0.3514) & 0.3642 & 0.3478 \\
{(}2/3, \phantom{1/}1] & 673 (0.7605) & 0.5037 & 0.5270 & 310 (0.3503) & 0.3917 & 0.4262 \\
overall & 885 (1.0000) & 0.5257 & 0.5174 & 885 (1.0000) & 0.2994 & 0.3071 \\
\midrule
\multicolumn{7}{c}{Ki for Target Carbonic anhydrase II (MAE)} \\
\midrule
\multirow{2}{*}{Bin} & \multicolumn{3}{c}{Randomized Split}     & \multicolumn{3}{c}{SAE (balanced) Split}  \\
\cmidrule(r){2-4} \cmidrule(r){5-7}
& Count (Ratio) & PharmHGT & SAM-DTA & Count (Ratio) & PharmHGT & SAM-DTA \\
\midrule
{[}0\phantom{/3}, 1/3] & \phantom{00}8 (0.0087) & 0.5879 & 0.6645 & 244 (0.2667) & 1.0450 & 1.0564 \\
{(}1/3, 2/3] & 201 (0.2197) & 0.6807 & 0.7009 & 342 (0.3738) & 0.7389 & 0.7572 \\
{(}2/3, \phantom{1/}1] & 706 (0.7716) & 0.5615 & 0.5426 & 329 (0.3596) & 0.6172 & 0.5813 \\
overall & 915 (1.0000) & 0.5879 & 0.5785 & 915 (1.0000) & 0.7768 & 0.7738 \\
\midrule
\multicolumn{7}{c}{Ki for Target Carbonic anhydrase II (R$^2$)} \\
\midrule
{[}0\phantom{/3}, 1/3] & \phantom{00}8 (0.0087) & 0.6739 & 0.4277 & 244 (0.2667) & -0.0885 & 0.0087 \\
{(}1/3, 2/3] & 201 (0.2197) & 0.5803 & 0.5742 & 342 (0.3738) & 0.4657 & 0.4690 \\
{(}2/3, \phantom{1/}1] & 706 (0.7716) & 0.5509 & 0.5932 & 329 (0.3596) & 0.4760 & 0.5346 \\
overall & 915 (1.0000) & 0.5684 & 0.5938 & 915 (1.0000) & 0.3776 & 0.4192 \\

\bottomrule
\end{tabular}
\end{table}

For the prediction method, as shown in Figure~\ref{fig:benchmark_EGFR}, we select five state-of-the-art and representative DTA prediction methods. Molecular Contrastive Learning of Representations (MolCLR) sees samples as atom-bond graphs and employs GCN and GIN to learn the molecular representations by contrastive pairs~\citep{wang2022molecular}. Sequence-agnostic model for drug-target binding affinity prediction (SAM-DTA), on contrast, takes as input the Simplified Molecular-Input Line-Entry System (SMILES) of samples and processes it using 1D-CNN with dilated parallel residual blocks~\citep{hu2023sam}.
SMILES is also utilized in FusionDTA but is processed by a unified LSTM model with linear attention mechanism \citep{yuan2022fusiondta}.
Finally, we include two transformer-based methods. One is ChemBERTa which takes as input SMILES of samples and builds a model with 12 attention heads and 6 layers \citep{chithrananda2020ChemBERTa}. The other is PharmHGT which leverages a unique pharmacophoric-constrained heterogeneous molecule graph and two various transformers to extract chemical properties and predict molecular attributes \citep{jiang2023pharmacophoric}. 

We also investigate the problem with other tasks and datasets. Specifically, for the task of IC50 prediction, we also perform experiments in the dataset of target BACE1 with a total of 4,636 samples, and we further extend the experiments to the task of Ki prediction for targets Carbonic anhydrase I and Carbonic anhydrase II, with 5,307 and 5,487 samples respectively. For all of these datasets, we apply the same preprocessing as that of target EGFR, except that taking the negative logarithm form is not applicable to Ki datasets. The results are collectively presented in Table~\ref{tab:benchmark}. Here we choose PharmHGT and SAM-DTA as the example methods for a detailed showcase, while the comprehensive collection of results can be found in the appendix.

In summary, extensive experiments demonstrate the generality of the imbalanced distribution of samples caused by randomized split and the consequent overshadowing of low-similarity samples. This problem will be analyzed and addressed in the following section.

\section{Similarity Aware Evaluation}

In this section, we will elaborate on the proposed Similarity Aware Evaluation~(SAE) which aims at obtaining a test set with the desired distribution. We will exemplify the method for a test set that is uniform at similarity-based bins (see Figure~\ref{fig:benchmark_EGFR} for 3 similarity-based bins), and then extend it to other desired distributions.

The split for a test set that is uniform across similarity-based bins can be formulated as a combinatorial optimization problem as follows. Given a dataset $X=\{x_i, i=1, 2, ..., N\}$, a pairwise similarity matrix $\{s_{ij}\in [0,1], s_{ii}=0, i=1, 2, ..., N; j=1, 2, ..., N\}$, a ratio $\alpha$, and $K$ bins with boundaries $\{b_k, k=0,1,2,...,K\}$, find a subset (test set) $X_\mathit{ts} \subset X, |X_\mathit{ts}| = \alpha N$, such that
\begin{align}
f(X_\mathit{ts}) &= \sum_{k=1}^K\frac{(o_k - \alpha N/K)^2}{\alpha N/K}\\
\intertext{is minimized, where}
o_k &= |\{x_i\in X_\mathit{ts}: b_{k-1}< r_i \le b_k \}|\\
r_i &= \max_{x_j\in X_\mathit{tr}}s_{ij}\\
X_\mathit{tr} &= X-X_\mathit{ts}
\end{align}
$X_\mathit{tr}$ denotes the training set, $r_i$ the similarity of $x_i$ to the training set, and $o_k$ the count for each of the $K$ bins. Note that the objective function $f$ is essentially the $\chi^2$ statistics in the Chi-Square ($\chi^2$) Test, where $o_k$ is the observed count in each bin and $\alpha N/K$ is expected. Note also that we specially set $s_{ii}=0$ in the pairwise similarity matrix. This has no effect on the problem itself but can avoid that $r_i$ falls trivially to 1 due to the maximum operation for the relaxed problem below.

The combination optimization problem is infeasible to solve for optimum. As a result, we relax it to a continuous optimization problem where samples are allowed to coexist in the training set and test set by the introduction of the weights $\{w_i \in [0,1], i=1,2, ..., N\}$ and by $|X_\mathit{ts}| = \alpha N$ we have constraints $\sum_i w_i = \alpha N$. Next, we have to deal with non-differentiable operations in the objective function $f$ including taking the maximum and counting in similarity-based bins. For the maximum function in the calculation of $r_i$, we approximate it by the LogSumExp operation with a hyper-parameter $\beta$,
\begin{align}
\label{align:beta}
r_i = \max_{x_j\in X_\mathit{tr}}s_{ij} = \max_j (1-w_j) s_{ij} \approx \frac{1}{\beta}\log\sum_j \exp\left(\beta(1-w_j) s_{ij}\right)
\end{align}

In terms of counting for similarity-based bins in the calculation of $o_k$, we approximate the discrete event of a sample falling into a specific bin by a continuous score which depends on how far $r_i$ of the sample deviates from the center of the bin. The score function is defined following the bell-shaped normal distribution with the center of the bin as the expectation and a tunable standard deviation. For a sample, scores across all bins are normalized. Specifically, denote $c_k = (b_{k-1} + b_k) / 2$ as the center of each bin, $\sigma_k$ as the tunable standard deviation, we have:
\begin{align}
o_k &= |\{x_i\in X_\mathit{ts}: b_{k-1}< r_i \le b_k \}|\\
       &= \sum_i w_i \mathbb{I}(b_{k-1}< r_i \le b_k)\\
       &\approx \sum_i w_i\frac{\frac{1}{\sqrt{2\pi}\sigma_k}\exp\left( -(r_i - c_k)^2/(2\sigma_k^2) \right)}{\sum_{k'} \frac{1}{\sqrt{2\pi}\sigma_{k'}}\exp\left( -(r_i - c_{k'})^2/(2\sigma_{k'}^2) \right)}
\end{align}
where $\mathbb{I}$ is the indicator function. In this paper, we set $\sigma_k = \sigma, k=1,2,..., K$.
Thus, we obtain the following expression:
\begin{align}
\label{align:sigma}
o_k \approx \sum_i w_i\frac{\exp\left( -(r_i - c_k)^2/(2\sigma^2) \right)}{\sum_{k'} \exp\left( -(r_i - c_{k'})^2/(2\sigma^2) \right)}
= \sum_i w_i\mathop{\mathit{softmax}}_k\left(-\frac{(r_i - c_k)^2}{2\sigma^2}\right)
\end{align}

\begin{figure}[t]
  \centering
  \includegraphics[width=0.95\textwidth]{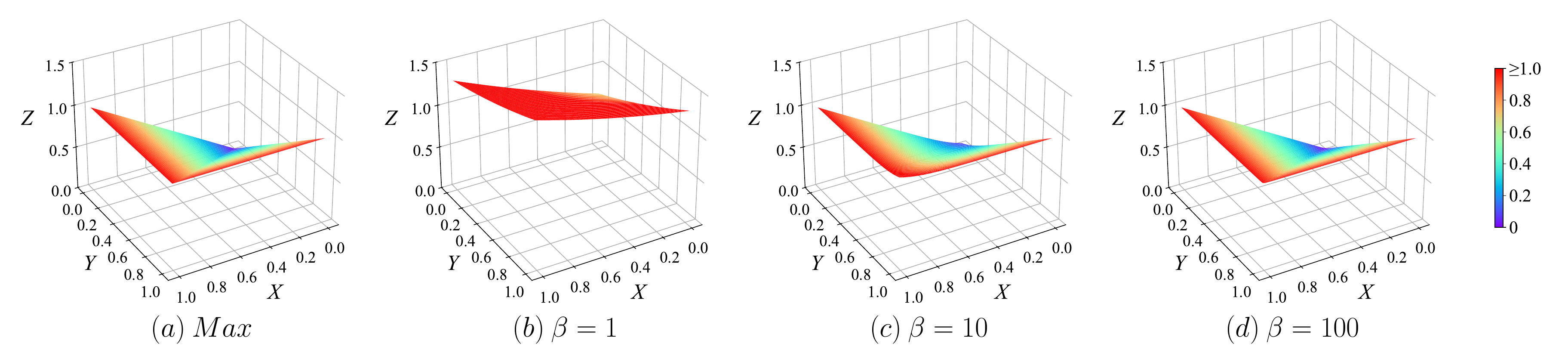}
  \caption{Impact of the hyper-parameter $\beta$ on the approximation of the maximum function in Eq.~\ref{align:beta}. To illustrate this impact, we consider a simplified scenario involving only two random variables: $X$ and $Y$. (a) $Z = Max(X, Y)$; (b-d) $Z = 1 / \beta \log (\exp(\beta X) + \exp(\beta Y))$. A larger value of $\beta$ results in a more accurate approximation, with $\beta = 100$ yielding an excellent result.}
  \label{fig:beta}
\end{figure}

\begin{figure}[t]
  \centering
  \includegraphics[width=0.9\textwidth]{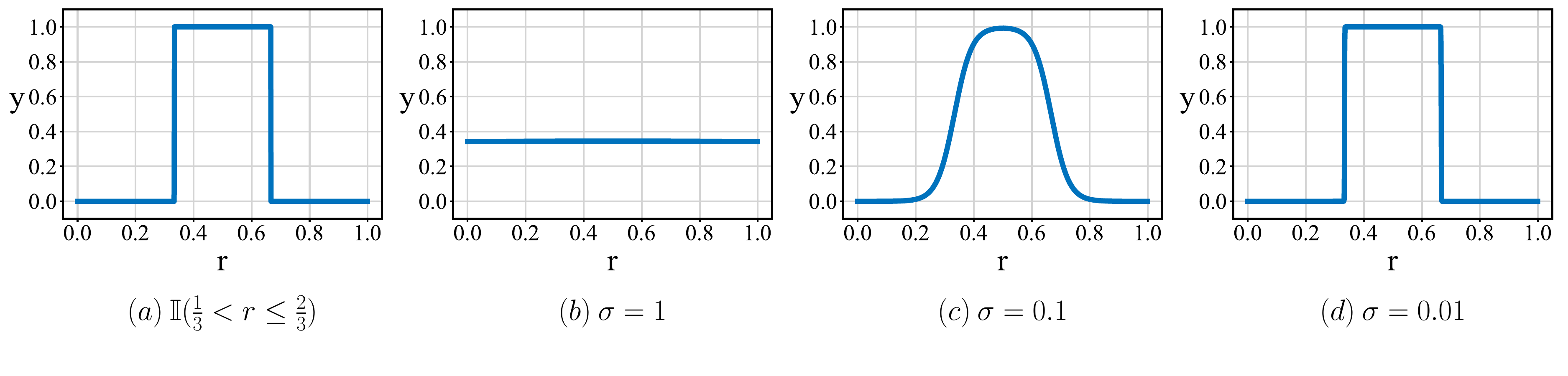}
  \vspace{-3mm}
  \caption{Influence of the hyper-parameter $\sigma$ in Eq.~\ref{align:sigma}. we analyze a specific case where $K = 3, b_k = k / 3, c_k = (2k - 1) / 6$. (a) $y = \mathbb{I}(b_1 < r \le b_2)$; (b-d) $y = \exp \left((-(r - c_2)^2 / (2\sigma^2)\right) / \sum_{k=1}^3 \exp \left(-(r - c_k)^2 / (2\sigma^2)\right)$. A decrease in the value of $\sigma$ leads to a more precise estimation, with $\sigma = 0.01$ producing an outstanding result.}
  \label{fig:sigma}
\end{figure}

Figure~\ref{fig:beta} and Figure~\ref{fig:sigma} illustrate the error induced by these two differentiable approximations, with respect to hyper-parameter $\beta$ and $\sigma$, respectively. In Figure~\ref{fig:beta} we compare $\beta$ between values of 1, 10, and 100 and plot the surface for a special case of maximum over two variables. It can be seen that a larger $\beta$ achieves a better approximation, but is also prone to overflow in practice. We use $\beta=100$ throughout the paper. For Figure~\ref{fig:sigma}, on the other hand, we show the comparison of $\sigma$ values between 1, 0.1, and 0.01 for an example case of the indicator of the second bin for a 3-bin setting $b_k = k/3$. The degree of approximation gets better when the value of $\sigma$ decreases, and is pretty well when $\sigma=0.01$. For the sake of flexibility, we set  $\sigma_k = 0.1 (b_k - b_{k - 1})$ in rest of the paper.

At the moment we seem to be ready to arrive at the approximated optimization function. However, in practice, we find that the approximation error induced by relaxing $w_i$ from $\{0, 1\}$ to $[0, 1]$ is not negligible. In fact, a considerable proportion of $w_i$ solved is neither near 0 nor 1. To address this issue, we are inspired by the concept of entropy, and propose to add a regularization term,
\begin{align}
l_\mathit{reg} &= -\lambda \sum_i \left( w_i\log(w_i) + (1-w_i)\log(1-w_i)\right)
\end{align}
where $\lambda$ is a hyper-parameter that balances between the objective function and the regularization term. Finally, we have the optimization problem,
\begin{align}
\label{align:prob}
\mathop{\mathit{minimize}}_{w_i} &\quad \sum_{k=1}^K\frac{(o_k - \alpha N/K)^2}{\alpha N/K} + l_\mathit{reg}\\
\mathit{subject\ to} &\quad \sum_i w_i  = \alpha N\\
&\quad 0\le w_i \le 1, i=1,2,...,N\\
\intertext{where}
o_k &= \sum_i w_i\mathop{\mathit{softmax}}_k\left(-\frac{(r_i - c_k)^2}{2\sigma^2}\right)\\
r_i &= \frac{1}{\beta}\log\sum_j \exp\left(\beta(1-w_j) s_{ij}\right)\\
l_\mathit{reg} &= -\lambda \sum_i \left( w_i\log(w_i) + (1-w_i)\log(1-w_i)\right)
\end{align}
Note that the optimization problem has no closed-form solution, and hence we have resorted to Lagrangian multipliers with numerical method implemented by PyTorch and Cooper.

For other desired distributions, one can modify the objective function $f$ in a straightforward way while the approximation tricks and regularization term can be retained, and the resulting optimization function can also be solved by Lagrangian multipliers with numerical method. Generally, if the expected count in each bin is $e_k, k=1,2,..., K$, the objective function can be readily modified as
\begin{align}
 \sum_{k=1}^K\frac{(o_k - e_k)^2}{e_k} + l_\mathit{reg}
\end{align}

\section{Experiments}

\subsection{Balanced split}
In Section~\ref{sec:prob_random}, we demonstrated that within the context of the randomized split, suboptimal performance on low-similarity samples does not significantly impact the overall performance, as they only occupy a negligible proportion.
To avoid disregarding samples with low similarity, we implemented a ``balanced split'' using similarity aware split strategy to achieve a uniformly distributed test set across various similarity bins ($[1/3, 2/3], (1/3, 2/3], (2/3, 1]$). Figure~\ref{fig:benchmark_EGFR} shows a comparison of randomized split and SAE (balanced) split at IC50 for EGFR. The randomized split strategy yielded a case in which 88\% of test samples have high similarity ($>2/3$) to the training set, while our split strategy guarantees a more evenly distributed range of similarities.
The evaluation at the SAE (balanced) split reveals that the performance of each model aligns with the respective similarity levels.
Hence, our SAE (balanced) split provides a more accurate representation of the performance.

Additional results at other tasks and datasets are delineated in Table~\ref{tab:benchmark}. Given the space constraint, we provide experimental results of two representative DTA prediction methods. Notably, analogous phenomena are observed across the remaining three datasets. The comprehensive collection of results can be found in the appendix.

\begin{figure}[ht]
  \centering
  \includegraphics[width=0.95\textwidth]{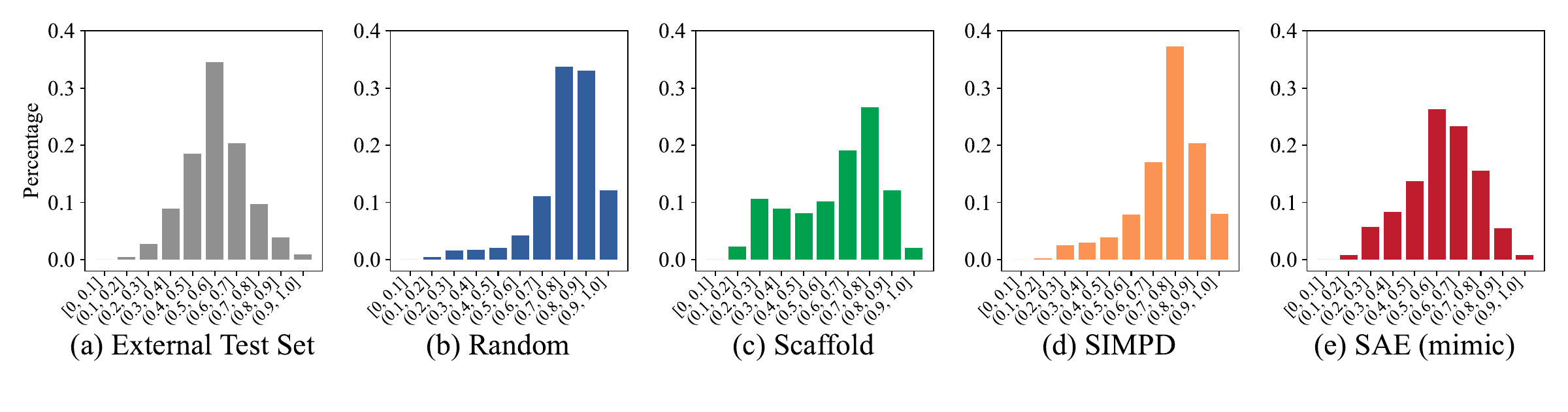}
  \caption{The similarity distribution of the internal test set across different split strategies. (b) Randomized split leads to a scenario where most internal test samples are highly similar to the training set. (c) Scaffold split produces a more balanced distribution. (d) SIMPD split yields a distribution similar to the random split. (e) Our SAE (mimic) split brings the internal test set's distribution closest to that of the external test set.}
  \label{fig:mimic_split_bar}
\end{figure}

\begin{figure}[ht]
  \centering
  \includegraphics[width=1.0\textwidth]{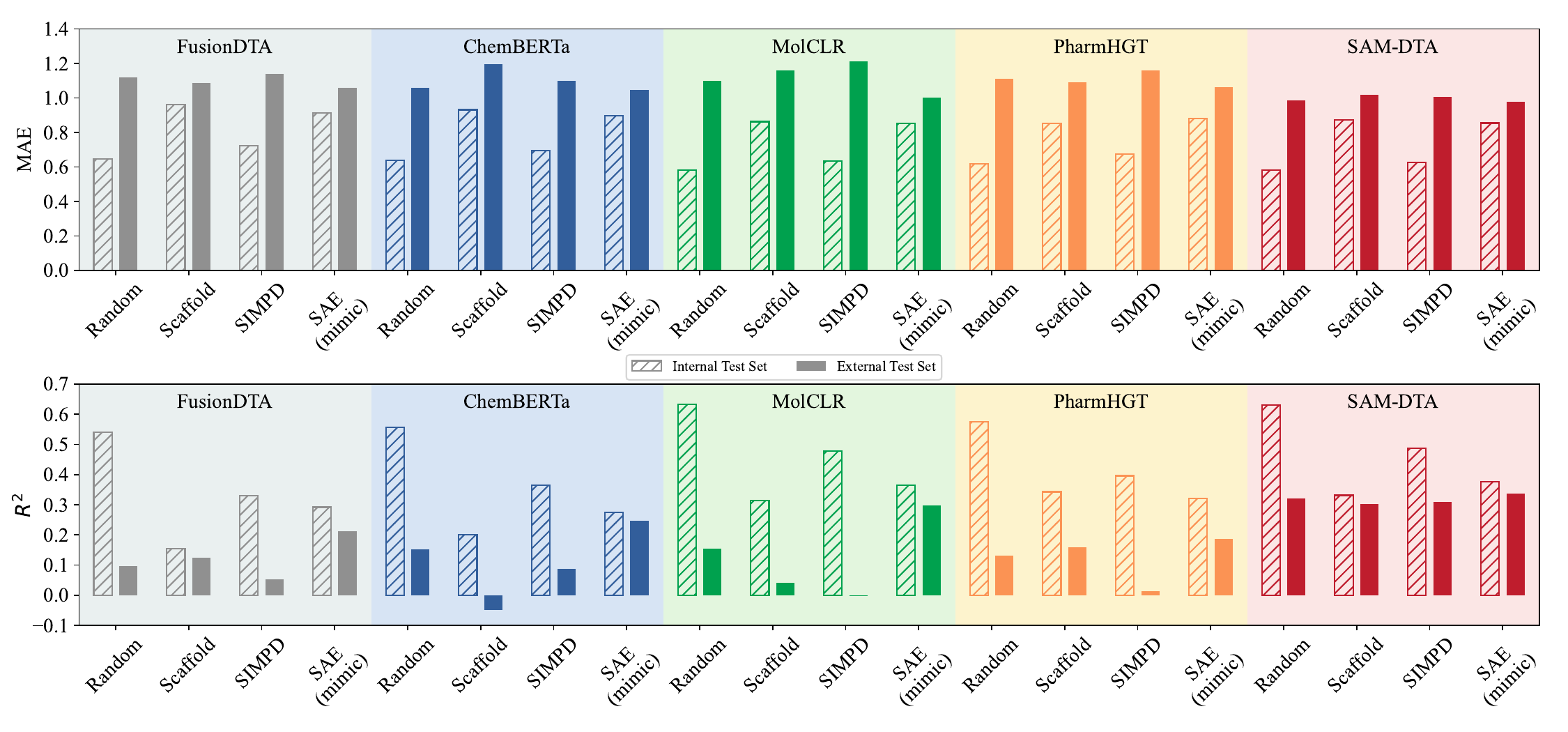}
  \caption{Comparison of the generalization ability of different split strategies at IC50 for EGFR across five DTA prediction methods.
  The external test set performance of the mimic split surpasses that of other split strategies.
  }
  \label{fig:mimic_split_score}
\end{figure}

\subsection{Mimic Split}

When prior knowledge about the external dataset—such as the distribution of similarity to existing samples—is available, we can construct an internal test set that mirrors this distribution. This approach enables optimal hyperparameter selection for the DTA prediction method, thereby enhancing performance on the external dataset.

We conducted experiments on the task of predicting IC50 values using the EGFR target dataset. Specifically, 70.2\% samples were procured from ChEMBL~\citep{zdrazil2024chembl}, while the remaining samples were obtained from PubChem~\citep{kim2023pubchem}, the US Patent and the scientific literature available in BindingDB~\citep{gilson2016bindingdb}.
For our analysis, we classified the ChEMBL-derived samples as internal data, while those obtained from the other sources as the external test set.
We first computed the similarity distribution between the external test set and the internal data, as shown in Figure~\ref{fig:mimic_split_bar} (a). Subsequently, we employed the Randomized split, Scaffold split, and SIMPD split~\citep{landrum2023simpd} to split the internal data into a training set and an internal test set with a ratio of 70\% and 30\%. The similarity distributions between the internal test set and the training set for these splits are depicted in Figure~\ref{fig:mimic_split_bar} (b-d), respectively. Finally, we utilized the proposed strategy to split the internal data, thereby mimicking the similarity distribution observed in the external test set. The results are illustrated in Figure~\ref{fig:mimic_split_bar} (e). We refer to this split strategy as ``mimic split''.

We searched for hyper-parameters such as optimizer, learning rate, batch size, and other relevant hyper-parameters. Details of the hyper-parameters for each method are provided in the appendix. 
Experimental results are shown in Figure~\ref{fig:mimic_split_score}, our SAE (mimic) split strategy consistently yields optimal hyper-parameter sets for all five DTA prediction methods. Among the various split strategies, the scores of our SAE (mimic) split on the internal test set are the most closely aligned with those on the external test set.

\begin{figure}[t]
  \centering
  \includegraphics[width=0.8\textwidth]{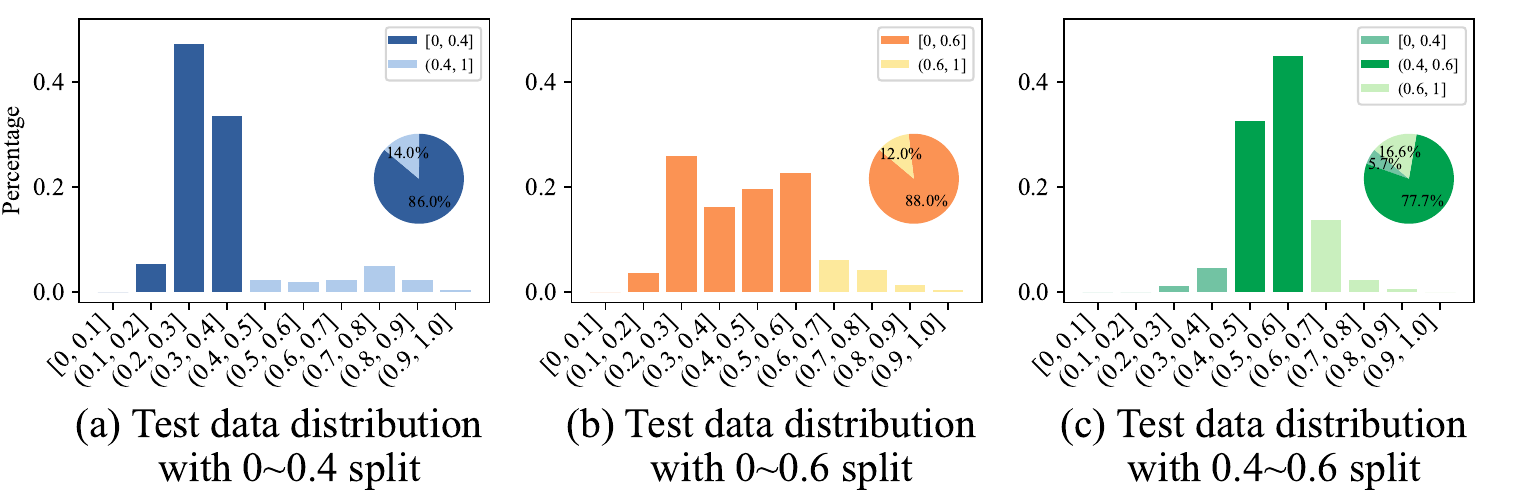}
  \caption{Other applications of our split strategy on the IC50 dataset of target EGFR. (a) 86\% of the test samples satisfied the desired distribution with a maximum similarity of 0.4, (b) 88\% of the test samples met the criteria for a maximum similarity of 0.6, and (c) 77.7\% of the test samples fulfilled the requirements for a similarity range between 0.4 and 0.6.}
  \label{fig:other_app_bar}
\end{figure}

\subsection{Other Applications}

Beyond achieving balanced splits, our strategy supports distributions with maximum similarities of 0.4 or 0.6 or a range between 0.4 and 0.6. In a 7:3 train-test split on the EGFR target's IC50 dataset (Figure~\ref{fig:other_app_bar}), SAE ensured 86\%, 88\%, and 77.7\% of test samples met the criteria, respectively. This underscores the flexibility of the strategy in accommodating diverse split needs.
Moreover, since SAE can flexibly achieve desired distributions by capturing the similarity between pairs of data samples, it can also be applied to QSAR scenarios, including ADMET prediction, drug design~\citep{de2022prediction, tropsha2024integrating}, as well as the prediction of protein-protein interactions (PPI)~\citep{sharma2021recent} and drug-drug interactions (DDI)~\citep{dmitriev2019prediction}.

\section{Related works}
When evaluating machine learning methods, it is essential to set aside a test set for benchmarking~\citep{wu2018moleculenet}. The similarity between the training set and the test set significantly influences the performance of these methods~\citep{sheridan2004similarity,cherkasov2014qsar,pahikkala2015toward,sieg2019need,nguyen2022mitigating,atas2023approach,harren2024modern}.
However, in the field of chemical data, imbalanced data distributions are an inherent and unavoidable challenge~\citep{harren2024modern,yang2020predicting,tossou2024real}. Therefore, it is crucial to design dataset split strategies that account for these imbalances and ensure meaningful evaluation of model performance~\citep{li2017structural, sheridan2022prediction}.
The commonly used random splitting method may fail to meet the requirements due to inherent data bias. A typical solution is to exclude all samples in the training set that are similar to those in the test set~\citep{li2021machine, scantlebury2023small, luo2024enhancing, luo2017network, wan2019neodti, atas2023approach}. Recently, several advanced split strategies have been proposed, including scaffold split~\citep{bemis1996properties, fang2022geometry,zhou2023uni,liu2024rethinking}, time split~\citep{guan20233d,stark2022equibind}, stratified split~\citep{wu2018moleculenet, chen2022meta}, physicochemical properties-based split \citep{kalemati2024dcgan}, cold-drug split~\citep{huang2021therapeutics}, and SIMPD ~\citep{landrum2023simpd}, among others.

\section{Conclusion}
In this paper, we show the generality of the imbalanced distribution of samples by randomized split and the consequent overshadowing of low-similarity samples.
To address the issue, we proposed a novel and flexible similarity aware split strategy for the test set to achieve a desired distribution like uniform discrete distribution, which can deliver a comprehensive evaluation for various drug-target binding affinity prediction algorithms.
Furthermore, we utilized the similarity aware split to create a ``mimic split'', splitting the training set and internal test set by replicating the distribution found in an external test set. Our mimic split consistently aids in selecting the optimal hyper-parameter across various deep-learning methods.
In the end, our split strategy enables the generation of distributions with minimum or maximum similarity constraints as required.

\section{Acknowledgement}
This work is partially supported by the Shanghai Artificial Intelligence Laboratory.

\bibliography{iclr2025_conference}
\bibliographystyle{iclr2025_conference}

\newpage
\appendix

\renewcommand{\thetable}{A\arabic{table}}
\renewcommand{\thefigure}{A\arabic{figure}}
\counterwithin{table}{section}
\counterwithin{figure}{section}

\section{Appendix}
\subsection{Related works on data splitting}
In molecular machine learning, including general QSAR tasks, the challenge of fair predictive evaluation has been a longstanding issue~\citep{cherkasov2014qsar}.
While randomized split remains the most commonly used strategy for data splitting, it is not always the optimal choice for evaluating machine learning methods. Consequently, various alternative split strategies have been developed to better evaluate the machine learning methods:

\begin{itemize}
    \item \textit{Time split}~\citep{sheridan2013time, stark2022equibind, guan20233d} is employed for datasets containing temporal information, where the model is trained on historical data and tested on more recent data. It may effectively replicate real-world scenarios, however, a significant number of datasets are devoid of time-specific information. In some situations, when the time span is too large or the data distribution changes significantly over time, the model may struggle to perform well on the test set.
    \item \textit{Scaffold split}~\citep{bemis1996properties} is a technique that splits the dataset based on the structural framework of each sample. It is often leveraged in situations involving out-of-distribution data to provide a measure of generalization capabilities~\citep{stanley2021fs, fang2022geometry, zhou2023uni, liu2024rethinking}. Because scaffold split does not enforce stratification during the partitioning process, it may result in class imbalance~\citep{yang2019analyzing}.
    \item \textit{Stratified split} - also called stratified random sampling - is a sampling method designed to ensure that each fold of a dataset maintains the same distribution of classes as the entire dataset. It achieves this by first dividing the data into different output strata based on class labels and then executing a random partition with the guaranteeing that the entire label range is encompassed within each set~\citep{krstajic2014cross, wu2018moleculenet, mathai2020validation, chen2022meta}.
    \item \textit{Cold-drug split}~\citep{huang2021therapeutics} is a method for dividing a dataset into multi-protein prediction tasks, where the dataset is split based on entity types, such as proteins, drugs, or DNAs. The process begins by randomly splitting the dataset into training, validation, and test sets based on one chosen entity type. Subsequently, all data samples associated with a specific entity are assigned to the same set to ensure no overlap across splits, ensuring that there is no overlap of the chosen entity type across the splits.
    \item \textit{SIMPD split}~\citep{landrum2023simpd} mimics temporal splits in scenarios where temporal information is not accessible. This approach was developed by observing and analyzing disparities observed between earlier and subsequent samples within the scope of medicinal chemistry projects. 
    \item \textit{Dissimilar split}~\citep{atas2023approach} divides a dataset into training and test sets by ensuring that samples in each set are dissimilar. This strategy prevents similar samples from appearing in both sets, thereby increasing the difficulty of predictions.
\end{itemize}

This challenge is closely intertwined with the broader problem of out-of-distribution (OOD) generalization~\citep{tossou2024real}, demonstrating its relevance far beyond the confines of individual tasks such as DTA prediction.
In fact, machine learning models tend to perform well when the training set shares a similar distribution with the test set~\citep{leonard2006selection, puzyn2011investigating, cherkasov2014qsar}.
However, previous split strategies often yield test sets with distributions that closely mimic the training set (as shown in Figure~\ref{fig:mimic_split_bar_revise}).
Such alignment between the training and test set distributions can lead to overly optimistic assessments of a model’s generalization ability, as it fails to account for scenarios where the model is applied to data with significantly different characteristics.
SAE provides an effective solution to this issue by enabling more precise control over data distributions through its ability to capture the similarities between data samples. This approach ensures greater adaptability to a wider range of scenarios.

\subsection{Discussion about the application of QSAR scenarios}
Quantitative Structure-Activity Relationship (QSAR) modeling is a widely used in silico approach for predicting the biological or chemical properties of molecules~\citep{de2022prediction}. Previous studies on QSAR~\citep{sheridan2004similarity} have shown that prediction accuracy is highly correlated with the similarity between the molecule being predicted and its closest neighbor in the training set. This observation is similar to patterns found in the DTA prediction task.  Therefore, our SAE method can also be extended to QSAR tasks.

For instance, \citet{krishnan2021accelerating} introduced a de novo drug design method that incorporates a pre-trained model alongside transfer learning to generate novel inhibitors targeting the human JAK2 protein. In this approach, transfer learning was utilized to capture the features of the target-related chemical space. If the characteristics of the target-related chemical space—particularly the distribution of the external dataset—are already well understood, our SAE can be applied to replicate this distribution during the splitting of training and test sets, thereby enhancing the overall performance.

Similarly, in the task of protein-protein interaction prediction, improper construction of the data split among training, validation, and test sets can lead to severe data leakage and overly optimistic results~\citep{li2022recent}. To address this issue, one proposed solution is to divide the test set into three distinct classes~\citep{park2012flaws}: C1, where test pairs consist of proteins that are both present in the training set; C2, where test pairs involve one protein present in the training set; and C3, where neither protein in the test pair is found in the training set.
Notably, the three classes can be viewed as specific cases of our SAE split strategy. Furthermore, the SAE approach can be flexibly applied to constructing test sets with varying levels of difficulty to more effectively evaluate the model's generalization.

\subsection{Time Complexity and Space Complexity Analysis}

Given the number of iterations $M$, the number of samples $N$, and the number of bins $K$, we analyze the time complexity of a single iteration in Eq.~\ref{align:prob}, which involves both forward and backward propagation. 
During forward propagation, computing $o_k$ involves $O(N \cdot K)$ operations, as it requires iterating over $N$ samples and $K$ bins, with softmax and exponential computations. The computation of $r_i$ is more expensive, requiring $O(N^2)$ operations due to the nested summation over $N$ samples. The regularization term $l_{reg}$ involves a simple summation over $N$, contributing $O(N)$ operations. Combining these, the time complexity of one forward propagation is dominated by the $O(N^2)$ and $O(N \cdot K)$ terms, resulting in $O(N^2 + N \cdot K)$, which simplifies to $O(N^2)$ because $K \ll N$.
For backward propagation, the computation of gradients with respect to $w_i$ involves similar operations, which results in the same complexity of $O(N^2)$. 
Additionally, the process of checking constraints involves $N + 1$ Lagrangian multipliers. The forward and backward propagation for this constraint-checking step each have a complexity of $O(N)$.
Combining all of these components, the time complexity of one iteration is $O(N^2)$, and the total time complexity of SAE across all iterations is $O(M \cdot N^2)$.
The overall space complexity of SAE is primarily determined by the storage requirements for $s_{ij}$ and the intermediate values needed for computing gradients from $r_i$ to $w_j$. As a result, the space complexity is $O(N^2)$.

Empirically, for the IC50 dataset of target EGFR which contains $N=4,361$ samples, the desired distribution is a uniform over bins $[1/3, 2/3], (1/3, 2/3], (2/3, 1]$. We set the number of iterations to $M=20,000$. On a single 3090 GPU, SAE completes the process in approximately 270 seconds, utilizing 2,410 MiB of GPU memory.

We would also like to emphasize that SAE is used in the model development stage and as a result, when the model is developed, it will no longer affect the efficiency for high-throughput inference. Note also that SAE needs only to be performed once for a fixed dataset, meaning that it can be reused by different models as long as they are developed on the same dataset. As a result, the time it takes may be overwhelmed by the time used by the heavy model development. When scaling to large datasets, it should be noted that almost all operations within one iteration are parallelable, and thus it will benefit significantly from more powerful GPU devices and distributed computation.

\begin{figure}[t]
  \centering
  \includegraphics[width=0.85\textwidth]{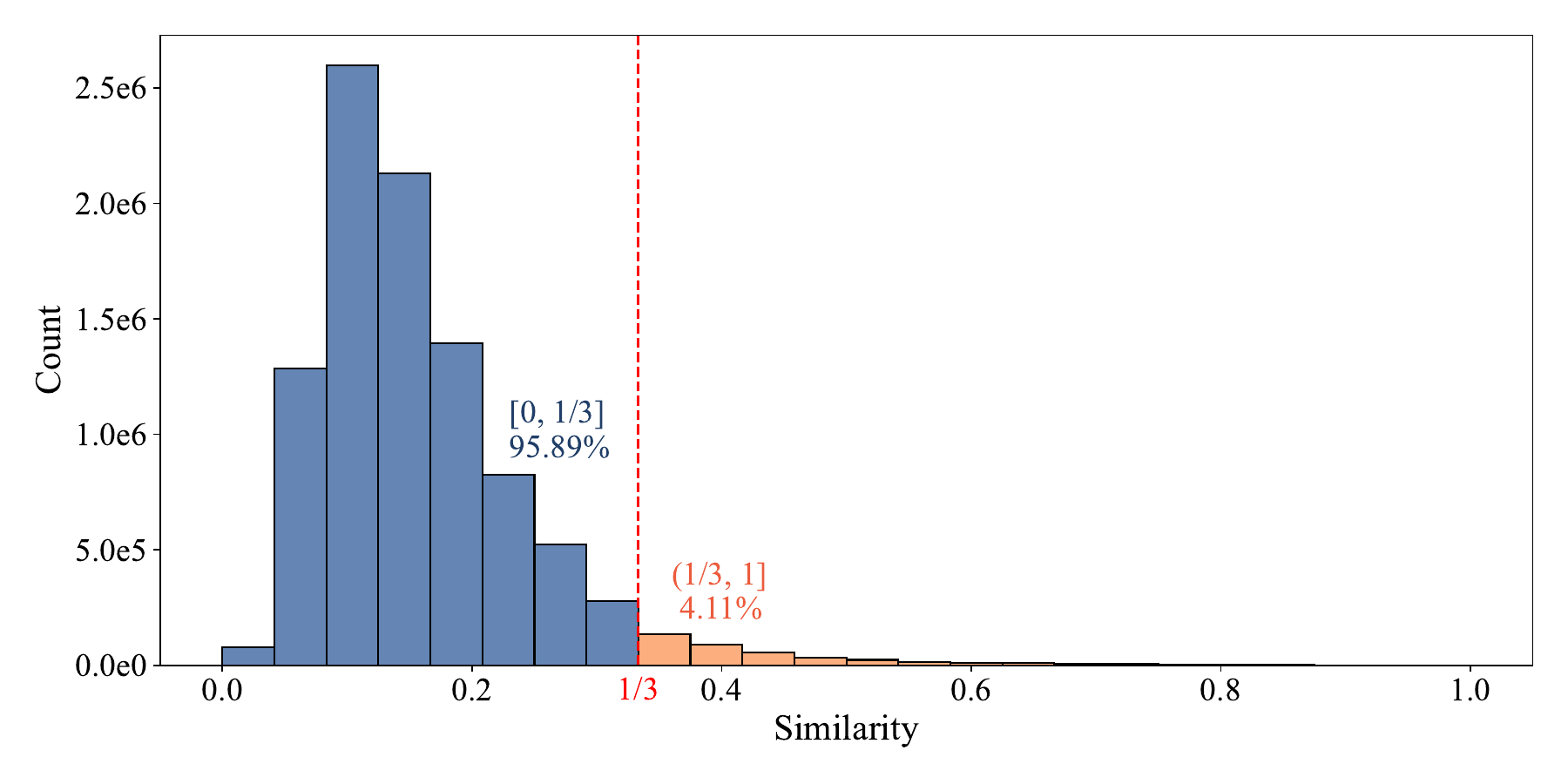}
  \caption{The similarity distribution of all sample pairs in the IC50 prediction task for the target EGFR. The dataset consists of 4,361 samples, resulting in a total of 9,506,980 pairwise similarity calculations. Among these, 95.89\% of the pairs exhibit similarities of no greater than 1/3, while only 4.11\% have similarities exceeding 1/3.}
  \label{fig:sim_hist}
\end{figure}

\begin{figure}[t]
  \centering
  \includegraphics[width=\textwidth]{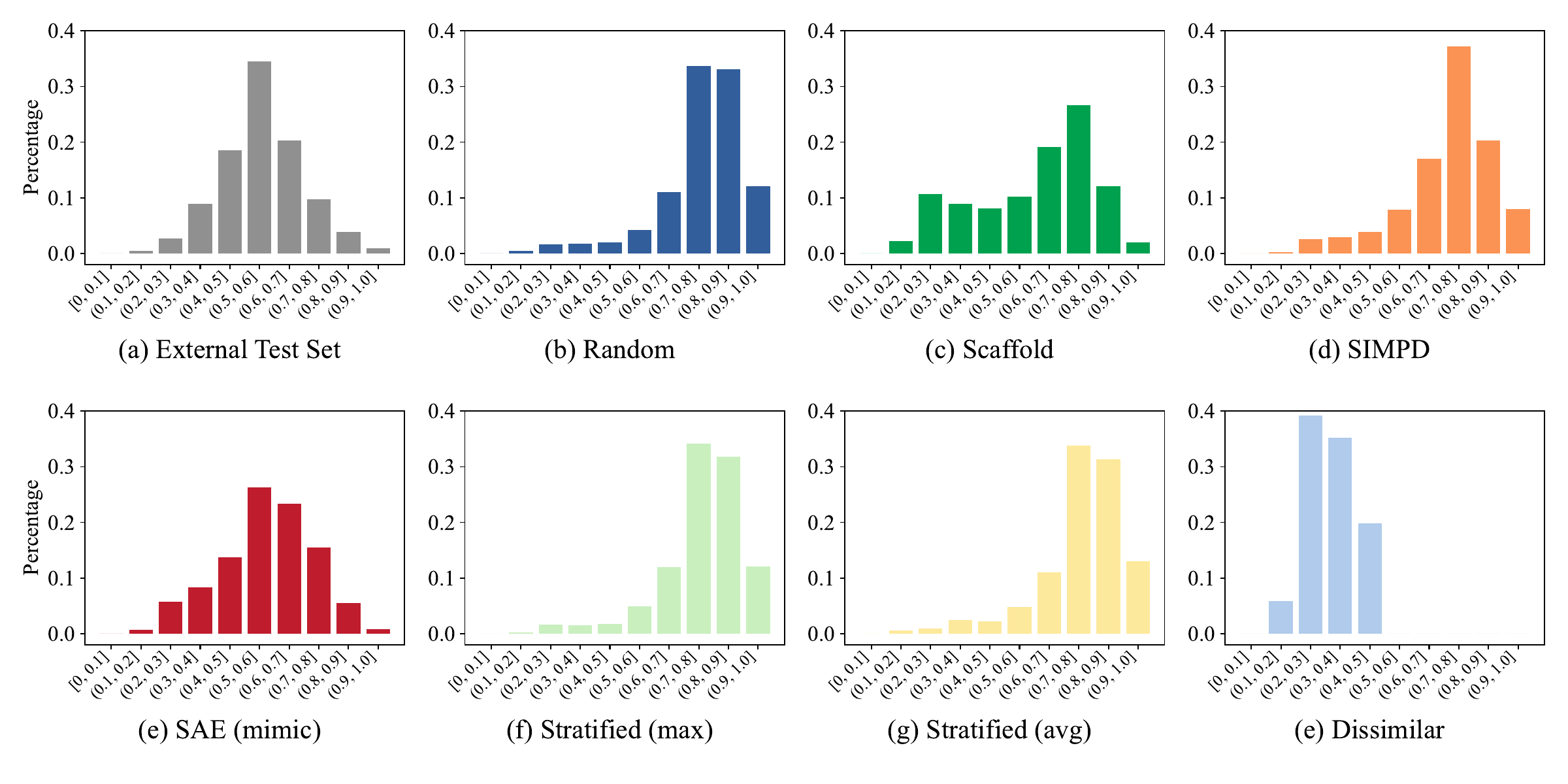}
  \caption{The similarity distribution of the internal test set across different split strategies. (b) Randomized split leads to a scenario where most internal test samples are highly similar to the training set. (c) Scaffold split produces a more balanced distribution. (d) SIMPD split yields a distribution similar to the random split. (e) Our SAE (mimic) split brings the internal test set's distribution closest to that of the external test set. (f) Stratified split based on the maximum similarity of each ligand to all others in the dataset. (g) Stratified split based on the average similarity of each ligand to all others in the dataset. (h) Dissimilar split guarantees that the similarity will remain below 0.5.}
  \label{fig:mimic_split_bar_revise}
\end{figure}

\subsection{Discussion on the application of SAE to large-scale datasets}
In pharmacompany (private) drug libraries for early drug discovery, there might be 200,000 to 10$^6$ samples~\citep{hughes2011principles}.
When scaling to these large datasets, a solution based on the sparse matrix is applicable. Specifically, Figure~\ref{fig:sim_hist} shows that the majority of pairwise similarities are low.
Suppose the desired distribution is uniform over bins $[1/3, 2/3], (1/3, 2/3], (2/3, 1]$ just as in the previous section, over 95\% entries in the similarity matrix are less than 1/3 and can be safely set to zero without the interference of the results. The time and space complexity can be significantly reduced in this way.
Given N = 10$^6$, the number of pairs with a similarity greater than 1/3 would be approximately $0.05 \cdot {N(N - 1)} / {2}$, which is about $2.5 \times 10^{10}$. We can store these similarities in a sparse format, represented as tuples (Index of sample A, Index of sample B, Similarity value). Each index can be encoded using 20 bits (sufficient to represent 2$^{20}$ = 1,048,576 positions), and the similarity value can be quantized into 4 bits~\citep{dettmers2024qlora}. Consequently, the total storage requirement can be calculated as:
\begin{align*}
    (20 bits + 20 bits + 4 bits) \times 2.5 \times 10^{10} = 5.5 Bytes \times 2.5 \times 10^{10} \approx 128 GB
\end{align*}
This size is manageable and can even be stored in memory.

\input{appendix_tables/mimic_split_detailed}
\input{appendix_tables/mimic_split_bins}

\subsection{Supplementary Experimental Results of Mimic Split}
For thorough comparison with other split strategies, we implemented stratified split~\citep{wu2018moleculenet, chen2022meta} and dissimilar split~\citep{atas2023approach} at IC50 for EGFR. For the stratified split, we first compute the pairwise similarities for the full dataset, resulting in a similarity matrix of size $N\times N$ (where $N$ is the number of samples in the dataset, with the diagonal values set to zero). Next, we calculate the maximum/average similarity for each row, yielding a similarity vector of size $N$, which represents the maximum/average similarity of each ligand to all others in the dataset. Finally, we divide the dataset into K bins based on the maximal/average similarity and perform random sampling within each bin to create the test set. We refer to the two variations of this stratified split strategy as "Stratified (max)," which uses the maximum similarity for binning, and "Stratified (avg)", which uses the average similarity.
The similarity distributions of the stratified split are shown in Figure~\ref{fig:mimic_split_bar_revise}~(f) and Figure~\ref{fig:mimic_split_bar_revise}~(g).
The distribution result of dissimilar split is shown in Figure~\ref{fig:mimic_split_bar_revise}~(h).
A comparison of the generalization ability of different split strategies is shown in Table~\ref{tab:mimic_revise}. SAE performs better than stratified sampling and dissimilar split with a clear margin.

We also analyze the performance of the different brackets in the external dataset. As is shown in Table~\ref{tab:mimic_bins_revise}, SAE improves performance in the low- and mid-similarity brackets, but not in the high-similarity one. We believe this is because the internal test set by SAE has more samples in the low- and mid-similarity brackets and thus performance in these brackets receives more attention compared with other split strategies.

\subsection{Comparison across different similarity measures and fingerprints}
As for comparison across different similarity measures and fingerprints, we conducted experiments on similarity measure choices including Tanimoto, Cosine, Sokal, and Dice, and fingerprint choices including Morgan (ECFP), RDKFP (RDKit), and Avalon. As shown by Table~\ref{cmp_sim_fp}, split results are less affected by similarity measure choices but more influenced by fingerprint choices. In all settings of similarity measures and fingerprints, SAE outperforms other approaches by achieving a split that is closer to the desired distribution (uniform distribution in this case).

\input{appendix_tables/cmp_sim_fp}

\input{appendix_tables/hyper_param}
\input{appendix_tables/random_FusionDTA_ChemBERTa}
\input{appendix_tables/random_MolCLR}
\input{appendix_tables/random_versus_SAE_FusionDTA_ChemBERTa}
\input{appendix_tables/random_versus_SAE_MolCLR}

\end{document}

%% file: math_commands.tex
\usepackage{amsmath,amsfonts,bm}

\def\eqref#1{equation~\ref{#1}}

\def\1{\bm{1}}

\DeclareMathAlphabet{\mathsfit}{\encodingdefault}{\sfdefault}{m}{sl}
\SetMathAlphabet{\mathsfit}{bold}{\encodingdefault}{\sfdefault}{bx}{n}



%% file: appendix_tables/mimic_split_detailed.tex
\begin{table}[t] 
\centering 
\caption{Comparison of the generalization ability of different split strategies at IC50 for EGFR across five DTA prediction methods.} 
\label{tab:mimic_revise}
\begin{tabular}{cccccc} 
\toprule 
\multirow{2}{*}{Split} & \multirow{2}{*}{Method} & Internal Test & Internal Test & External Test & External Test \\
& & MAE & R$^2$ & MAE & R$^2$ \\
\midrule

\multirow{5}{*}{Random} & FusionDTA & 0.6445 & 0.5399 & 1.1198 & 0.0957 \\
& ChemBERTa & 0.6389 & 0.5574 & 1.0600 & 0.1517 \\
& MolCLR & 0.5830 & 0.6318 & 1.0976 & 0.1546 \\
& PharmHGT & 0.6166 & 0.5739 & 1.1107 & 0.1323 \\
& SAM-DTA & 0.5805 & 0.6301 & 0.9863 & 0.3206 \\

\midrule

\multirow{5}{*}{Scaffold} & FusionDTA & 0.9626 & 0.1544 & 1.0863 & 0.1243 \\
& ChemBERTa & 0.9314 & 0.1997 & 1.1972 & -0.0491 \\
& MolCLR & 0.8627 & 0.3145 & 1.1585 & 0.0405 \\
& PharmHGT & 0.8537 & 0.3427 & 1.0930 & 0.1594 \\
& SAM-DTA & 0.8725 & 0.3311 & 1.0187 & 0.3034 \\

\midrule

\multirow{5}{*}{SIMPD} & FusionDTA & 0.7215 & 0.3292 & 1.1417 & 0.0528 \\
& ChemBERTa & 0.6954 & 0.3642 & 1.1010 & 0.0878 \\
& MolCLR & 0.6334 & 0.4775 & 1.2131 & -0.0016 \\
& PharmHGT & 0.6742 & 0.3958 & 1.1588 & 0.0133 \\
& SAM-DTA & 0.6271 & 0.4867 & 1.0058 & 0.3083 \\
\midrule

\multirow{5}{*}{Stratified (max)} & FusionDTA & 0.6753 & 0.5346 & 1.0886 & 0.1517 \\
& ChemBERTa & 0.6752 & 0.5384 & 1.1504 & 0.0223 \\
& MolCLR & 0.5968 & 0.6504 & 1.0917 & 0.1207 \\
& PharmHGT & 0.6092 & 0.6302 & 1.0694 & 0.1811 \\
& SAM-DTA & 0.6019 & 0.6404 & 1.0345 & 0.2722 \\
\midrule

\multirow{5}{*}{Stratified (avg)} & FusionDTA & 0.6490 & 0.5713 & 1.0957 & 0.1206 \\
& ChemBERTa & 0.6724 & 0.5191 & 1.1258 & 0.0896 \\ 
& MolCLR & 0.5939 & 0.6368 & 1.1556 & 0.1019 \\
& PharmHGT & 0.6103 & 0.5895 & 1.0938 & 0.1667 \\
& SAM-DTA & 0.6099 & 0.6159 & 0.9946 & 0.3345 \\
\midrule

\multirow{5}{*}{Dissimilar} & FusionDTA & 0.9425 & -0.1256 & 1.2788 & -0.1063 \\
& ChemBERTa & 0.8927 & -0.0139 & 1.6402 & -0.5971 \\
& MolCLR & 0.8462 & 0.0592 & 1.3355 & -0.1366 \\
& PharmHGT & 0.9011 & -0.0029 & 1.6006 & -0.5237 \\
& SAM-DTA & 0.9239 & -0.0845 & 1.2140 & -0.0039 \\
\midrule

\multirow{5}{*}{SAE (mimic)} & FusionDTA & 0.9130 & 0.2919 & 1.0605 & 0.2122 \\
& ChemBERTa & 0.8976 & 0.2736 & 1.0452 & 0.2477 \\
& MolCLR & 0.8536 & 0.3653 & 1.0002 & 0.2981 \\
& PharmHGT & 0.8826 & 0.3200 & 1.0609 & 0.1861 \\
& SAM-DTA & 0.8545 & 0.3770 & 0.9773 & 0.3367 \\

\bottomrule
\end{tabular}
\end{table}

%% file: appendix_tables/mimic_split_bins.tex
\begin{table}[t] 
\centering 
\caption{Detailed comparison of the generalization ability of different split strategies at IC50 for EGFR across five DTA prediction methods.}
\label{tab:mimic_bins_revise}
\begin{tabular}{c@{\hspace{1mm}}c@{\hspace{2mm}}c@{\hspace{2mm}}c@{\hspace{1mm}}c@{\hspace{1mm}}c@{\hspace{1mm}}c@{\hspace{1mm}}c} 
\toprule
\multicolumn{8}{c}{Extrenal Test MAE} \\
\midrule
Bin & Count & Split & FusionDTA & ChemBERTa & MolCLR & PharmHGT & SAM-DTA \\
\midrule

\multirow{6}{*}{[0, 1/3]} & \multirow{6}{*}{120} &Random & 1.2442 & 1.1167 & 1.2385 & 1.3261 & 1.1343 \\
 & & Scaffold & 1.2293 & 1.2242 & 1.1630 & 1.1840 & 1.1363 \\
 & & SIMPD & 1.4297 & 1.3223 & 1.7376 & 1.3382 & 1.1179 \\
 & & Stratified (max) & 1.2311 & 1.1663 & 1.2414 & 1.2179 & 1.1747 \\
 & & Stratified (avg) & 1.5371 & 1.4478 & 1.6005 & 1.3962 & 1.3666 \\
 & & Dissimilar & 1.4134 & 1.2363 & 1.1912 & 1.2424 & 1.4161 \\
 & & SAE (mimic) & 1.0626 & 1.0315 & 1.1848 & 1.1082 & 1.0435 \\
\midrule
\multirow{6}{*}{(1/3, 2/3]} & \multirow{6}{*}{1026} &Random & 1.1416 & 1.0874 & 1.0983 & 1.0989 & 0.9879 \\
 & & Scaffold & 1.0729 & 1.2273 & 1.2021 & 1.0891 & 1.0209 \\
 & & SIMPD & 1.1545 & 1.1272 & 1.2134 & 1.1614 & 1.0158 \\
 & & Stratified (max) & 1.0756 & 1.1611 & 1.0927 & 1.0677 & 1.0317 \\
 & & Stratified (avg) & 1.0815 & 1.1450 & 1.1384 & 1.0835 & 0.9808 \\
 & & Dissimilar & 1.3106 & 1.6954 & 1.3676 & 1.6504 & 1.2513 \\
 & & SAE (mimic) & 1.0531 & 1.0594 & 0.9979 & 1.0606 & 0.9743 \\
\midrule
\multirow{6}{*}{(2/3, 1]} & \multirow{6}{*}{186} &Random & 0.9559 & 0.9060 & 1.0208 & 1.0516 & 0.9015 \\
 & & Scaffold & 1.0604 & 1.0346 & 0.9449 & 1.0543 & 0.9335 \\
 & & SIMPD & 0.9741 & 0.9022 & 1.0066 & 1.0775 & 0.9191 \\
 & & Stratified (max) & 1.0718 & 1.0883 & 1.0017 & 0.9933 & 0.9683 \\
 & & Stratified (avg) & 0.9694 & 0.9047 & 1.0415 & 1.0102 & 0.8967 \\
 & & Dissimilar & 1.0201 & 1.6720 & 1.2858 & 1.6248 & 0.8773 \\
 & & SAE (mimic) & 1.1003 & 0.9762 & 0.8938 & 1.0322 & 0.9511 \\

\midrule
\multicolumn{8}{c}{External Test R$^2$} \\
\midrule
Bin & Count & Split & FusionDTA & ChemBERTa & MolCLR & PharmHGT & SAM-DTA \\
\midrule

\multirow{6}{*}{[0, 1/3]} & \multirow{6}{*}{120} &Random & -0.5461 & -0.5827 & -0.5009 & -0.6605 & -0.2855 \\
 & & Scaffold & -0.1335 & -0.3903 & -0.1887 & -0.1096 & -0.0121 \\
 & & SIMPD & -0.8717 & -0.7917 & -1.3400 & -0.6245 & -0.1181 \\
 & & Stratified (max) & -0.4278 & -0.4195 & -0.7345 & -0.3873 & -0.3106 \\
 & & Stratified (avg) & -0.8887 & -1.0024 & -1.1185 & -0.5790 & -0.3962 \\
 & & Dissimilar & -0.4209 & -0.0067 & 0.0122 & -0.0173 & -0.4122 \\
 & & SAE (mimic) & -0.0736 & -0.1140 & -0.3105 & -0.2592 & -0.0783 \\
\midrule
\multirow{6}{*}{(1/3, 2/3]} & \multirow{6}{*}{1026} &Random & 0.0351 & 0.1066 & 0.1433 & 0.1005 & 0.2950 \\
 & & Scaffold & 0.0987 & -0.1211 & -0.0502 & 0.1355 & 0.2824 \\
 & & SIMPD & 0.0078 & 0.0332 & -0.0169 & -0.0337 & 0.2780 \\
 & & Stratified (max) & 0.1305 & -0.0321 & 0.1016 & 0.1530 & 0.2633 \\
 & & Stratified (avg) & 0.1017 & 0.0476 & 0.1156 & 0.1430 & 0.3188 \\
 & & Dissimilar & -0.2307 & -0.8279 & -0.2834 & -0.7374 & -0.1244 \\
 & & SAE (mimic) & 0.1791 & 0.1848 & 0.2632 & 0.1237 & 0.3020 \\
\midrule
\multirow{6}{*}{(2/3, 1]} & \multirow{6}{*}{186} &Random & 0.0869 & 0.1390 & -0.0844 & 0.0577 & 0.2594 \\
 & & Scaffold & -0.0967 & -0.0351 & 0.1475 & -0.0412 & 0.2110 \\
 & & SIMPD & 0.1642 & 0.2476 & 0.0842 & 0.0139 & 0.2880 \\
 & & Stratified (max) & 0.0272 & -0.0595 & 0.1063 & 0.1111 & 0.1550 \\
 & & Stratified (avg) & 0.1114 & 0.2234 & 0.0081 & 0.0860 & 0.3187 \\
 & & Dissimilar & 0.0288 & -1.0679 & -0.3223 & -0.9576 & 0.2743 \\
 & & SAE (mimic) & -0.2135 & 0.1123 & 0.1807 & 0.0350 & 0.1359 \\

\bottomrule
\end{tabular}
\end{table}

%% file: appendix_tables/cmp_sim_fp.tex
\begin{sidewaystable}[h] 
\centering 

\caption{Comparison across different similarity measures and fingerprints, the desired distribution is a uniform distribution across the bins [0, 1/3], (1/3, 2/3], and (2/3, 1]. The numbers in this table are presented in the form [Sample counts in the first bin, Sample counts in the second bin, Sample counts in the third bin].}
\label{cmp_sim_fp}
\begin{tabular}{ccccccccc} 
\toprule
Similarity Measure & Fingerprint & SAE (balanced) & Random & Scaffold & SIMPD & Stratified (max) & Stratified (avg) \\
\midrule

\multirow{3}{*}{Cosine} & Morgan & 145, 436, 292 & \phantom{00}0, \phantom{0}33, 840   & \phantom{00}6, 159, 708   & \phantom{0}16, 657, 200  & \phantom{00}0, \phantom{0}32, 841   & \phantom{00}0, \phantom{0}27, 846   \\
& RDKFP  & \phantom{0}18, 426, 429  & \phantom{00}0, \phantom{0}17, 856   & \phantom{00}1, \phantom{0}78, 794    & \phantom{00}1, 124, 748   & \phantom{00}0, \phantom{0}13, 860   & \phantom{00}1, \phantom{0}14, 858   \\
& Avalon & \phantom{00}9, 429, 435   & \phantom{00}0, \phantom{00}7, 866    & \phantom{00}0, \phantom{0}21, 852    & \phantom{00}0, \phantom{0}16, 857    & \phantom{00}0, \phantom{00}6, 867    & \phantom{00}0, \phantom{00}6, 867    \\
\midrule
\multirow{3}{*}{Sokal} & Morgan & 292, 289, 292 & \phantom{0}33, 398, 442 & 172, 510, 191 & 689, 126, \phantom{0}58  & \phantom{0}34, 423, 416 & \phantom{0}29, 416, 428 \\
& RDKFP  & 291, 291, 291 & \phantom{0}19, \phantom{0}80, 774  & \phantom{0}85, 236, 552  & 135, 624, 114 & \phantom{0}14, \phantom{0}82, 777  & \phantom{0}15, \phantom{0}74, 784  \\
& Avalon & 291, 291, 291 & \phantom{00}7, \phantom{0}63, 803   & \phantom{0}26, 275, 572  & \phantom{0}23, 629, 221  & \phantom{00}8, \phantom{0}76, 789   & \phantom{00}6, \phantom{0}73, 794   \\

\midrule

\multirow{3}{*}{Dice}& Morgan & 182, 378, 313 & \phantom{00}0, \phantom{0}33, 840   & \phantom{00}9, 163, 701   & \phantom{0}17, 672, 184  & \phantom{00}0, \phantom{0}34, 839   & \phantom{00}2, \phantom{0}27, 844   \\
& RDKFP  & \phantom{0}60, 463, 350  & \phantom{00}0, \phantom{0}19, 854   & \phantom{00}2, \phantom{0}83, 788    & \phantom{00}1, 134, 738   & \phantom{00}0, \phantom{0}14, 859   & \phantom{00}1, \phantom{0}14, 858   \\
& Avalon & \phantom{0}32, 416, 425  & \phantom{00}0, \phantom{00}7, 866    & \phantom{00}0, \phantom{0}26, 847    & \phantom{00}0, \phantom{0}23, 850    & \phantom{00}0, \phantom{00}8, 865    & \phantom{00}0, \phantom{00}6, 867    \\

\midrule

\multirow{3}{*}{Tanimoto} & Morgan & 290, 299, 284 & \phantom{0}16, \phantom{0}98, 759  & \phantom{0}80, 273, 520  & 228, 547, \phantom{0}98  & \phantom{0}12, \phantom{0}99, 762  & \phantom{0}13, \phantom{0}99, 761  \\
& RDKFP  & 289, 292, 292 & \phantom{00}8, \phantom{0}34, 831   & \phantom{0}15, 184, 674  & \phantom{00}2, 635, 236   & \phantom{00}1, \phantom{0}39, 833   & \phantom{00}4, \phantom{0}33, 836   \\
& Avalon & 220, 325, 328 & \phantom{00}0, \phantom{0}28, 845   & \phantom{00}2, 154, 717   & \phantom{00}1, 324, 548   & \phantom{00}0, \phantom{0}30, 843   & \phantom{00}1, \phantom{0}28, 844   \\

\bottomrule
\end{tabular}

\end{sidewaystable}

%% file: appendix_tables/hyper_param.tex
\begin{table}[h] 
\centering 
\caption{Hyper-parameters used in the mimic split experiment for each method, the following search options are derived from the default parameter settings of each method.}
\begin{tabular}{ccc} 
\toprule
Method & Hyper-parameter & Options \\
\midrule
\multirow{3}{*}{SAM-DTA} & Optimizer & [Adam, SGD] \\
    & Learning rate & [1e-3, 1e-4, 1e-5] \\
    & Batch size & [10, 32, 64] \\
\midrule

\multirow{4}{*}{MolCLR} & Optimizer & [Adam, SGD] \\
    & Learning rate & \multirow{2}{*}{[(1e-3, 5e-3), (1e-4, 5e-4), (1e-5, 5e-5)]} \\
    & (prediction head, GNN encoder) & \\
    & Dropout ratio & [0.3, 0.5] \\
    & Readout pooling & [Mean, Max, Add] \\
\midrule

\multirow{4}{*}{FusionDTA} & Optimizer & [Adam, SGD] \\
    & Learning rate & [1e-2, 1e-3, 1e-4] \\
    & Batch size & [128, 256] \\
    & Loss function & [L1, MSE] \\
\midrule

\multirow{4}{*}{PharmHGT} &  Optimizer & [Adam, SGD] \\
    & Learning rate & [1e-2, 1e-3, 1e-4] \\
    & Activation function & [Sigmoid, ReLU] \\
    & Loss function & [RMSE, MAE] \\
\midrule

\multirow{3}{*}{ChemBERTa} & Optimizer & [AdamW, Adafactor] \\
    & Learning rate & [4e-3, 4e-4, 4e-5, 4e-6] \\
    & Batch size & [4, 8, 16] \\

\bottomrule
\end{tabular}
\end{table}

%% file: appendix_tables/random_FusionDTA_ChemBERTa.tex
\begin{table}[t]
\centering
\caption{Variations of $\mathit{SimilarityToTrainingSet}$ related to feature extraction, similarity measure, aggregation functions, and performance metrics. The methods for the detailed showcase are FusionDTA and ChemBERTa.}
\begin{tabular}{c@{\hspace{2mm}}c@{\hspace{2mm}}c@{\hspace{2mm}}c@{\hspace{2mm}}c@{\hspace{2mm}}c@{\hspace{2mm}}c}
\toprule

\multicolumn{7}{c}{Randomized Split (MAE)} \\
\midrule
\multirow{2}{*}{Bin} & \multicolumn{3}{c}{Feature: RDKit fingerprint} & \multicolumn{3}{c}{Feature:  Avalon fingerprint} \\
\cmidrule(r){2-4} \cmidrule(r){5-7}
& Count (Ratio) & FusionDTA & ChemBERTa & Count (Ratio) & FusionDTA & ChemBERTa \\
\midrule
{[}0\phantom{/3}, 1/3] & \phantom{00}8 (0.0092) & 1.4442 & 1.4679 & \phantom{00}0 (0.0000) & - & - \\
{(}1/3, 2/3] & \phantom{0}34 (0.0389) & 1.1294 & 1.1306 & \phantom{0}28 (0.0321) & 1.3299 & 1.2340 \\
{(}2/3, \phantom{1/}1] & 831 (0.9519) & 0.6407 & 0.6546 & 845 (0.9679) & 0.6451 & 0.6623 \\
overall & 873 (1.0000) & 0.6671 & 0.6806 & 873 (1.0000) & 0.6671 & 0.6806 \\
\midrule
 & \multicolumn{3}{c}{SimilarityMeasure: Sokal similarity} & \multicolumn{3}{c}{SimilarityMeasure: Dice coefficient} \\
\midrule
{[}0\phantom{/3}, 1/3] & \phantom{0}33 (0.0378) & 1.2751 & 1.2711 & \phantom{00}0 (0.0000) & - & - \\
{(}1/3, 2/3] & 398 (0.4559) & 0.7366 & 0.7234 & \phantom{0}33 (0.0378) & 1.2751 & 1.2711 \\
{(}2/3, \phantom{1/}1] & 442 (0.5063) & 0.5591 & 0.5980 & 840 (0.9622) & 0.6432 & 0.6574 \\
overall & 873 (1.0000) & 0.6671 & 0.6806 & 873 (1.0000) & 0.6671 & 0.6806 \\
\midrule
 & \multicolumn{3}{c}{Aggregation: Top-3} & \multicolumn{3}{c}{Aggregation: Top-5} \\
\midrule
{[}0\phantom{/3}, 1/3] & \phantom{0}17 (0.0195) & 1.1567 & 1.3484 & \phantom{0}24 (0.0275) & 1.3682 & 1.3330 \\
{(}1/3, 2/3] & 171 (0.1959) & 0.8839 & 0.8722 & 240 (0.2749) & 0.7861 & 0.8094 \\
{(}2/3, \phantom{1/}1] & 685 (0.7847) & 0.6008 & 0.6162 & 609 (0.6976) & 0.5926 & 0.6041 \\
overall & 873 (1.0000) & 0.6671 & 0.6806 & 873 (1.0000) & 0.6671 & 0.6806 \\
\midrule
\multicolumn{7}{c}{Randomized Split (R$^2$)} \\
\midrule
\multirow{2}{*}{Bin} & \multicolumn{3}{c}{Feature: RDKit fingerprint} & \multicolumn{3}{c}{Feature:  Avalon fingerprint} \\
\cmidrule(r){2-4} \cmidrule(r){5-7}
& Count (Ratio) & FusionDTA & ChemBERTa & Count (Ratio) & FusionDTA & ChemBERTa \\
\midrule
{[}0\phantom{/3}, 1/3] & \phantom{00}8 (0.0092) & 0.2319 & 0.2536 & \phantom{00}0 (0.0000) & - & - \\
{(}1/3, 2/3] & \phantom{0}34 (0.0389) & 0.2253 & 0.1829 & \phantom{0}28 (0.0321) & 0.2166 & 0.2573 \\
{(}2/3, \phantom{1/}1] & 831 (0.9519) & 0.5899 & 0.5796 & 845 (0.9679) & 0.5845 & 0.5698 \\
overall & 873 (1.0000) & 0.5697 & 0.5585 & 873 (1.0000) & 0.5697 & 0.5585 \\
\midrule
 & \multicolumn{3}{c}{SimilarityMeasure: Sokal similarity} & \multicolumn{3}{c}{SimilarityMeasure: Dice coefficient} \\
\midrule
{[}0\phantom{/3}, 1/3] & \phantom{0}33 (0.0378) & 0.0871 & 0.0615 & \phantom{00}0 (0.0000) & - & - \\
{(}1/3, 2/3] & 398 (0.4559) & 0.5068 & 0.5186 & \phantom{0}33 (0.0378) & 0.0871 & 0.0615 \\
{(}2/3, \phantom{1/}1] & 442 (0.5063) & 0.6469 & 0.6117 & 840 (0.9622) & 0.5898 & 0.5793 \\
overall & 873 (1.0000) & 0.5697 & 0.5585 & 873 (1.0000) & 0.5697 & 0.5585 \\
\midrule
 & \multicolumn{3}{c}{Aggregation: Top-3} & \multicolumn{3}{c}{Aggregation: Top-5} \\
\midrule
{[}0\phantom{/3}, 1/3] & \phantom{0}17 (0.0195) & -0.0685 & -0.3530 & \phantom{0}24 (0.0275) & -0.0237 & -0.0345 \\
{(}1/3, 2/3] & 171 (0.1959) & 0.4390 & 0.4408 & 240 (0.2749) & 0.4881 & 0.4647 \\
{(}2/3, \phantom{1/}1] & 685 (0.7847) & 0.5930 & 0.5837 & 609 (0.6976) & 0.5965 & 0.5903 \\
overall & 873 (1.0000) & 0.5697 & 0.5585 & 873 (1.0000) & 0.5697 & 0.5585 \\

\bottomrule 
\end{tabular}
\end{table}

%% file: appendix_tables/random_MolCLR.tex
\begin{table}[t]
\centering
\caption{Variations of $\mathit{SimilarityToTrainingSet}$ related to feature extraction, similarity measure, aggregation functions, and performance metrics. The method for the detailed showcase is MolCLR.}
\begin{tabular}{ccccc}
\toprule

\multicolumn{5}{c}{Randomized Split (MAE)} \\
\midrule
\multirow{2}{*}{Bin} & \multicolumn{2}{c}{Feature: RDKit fingerprint} & \multicolumn{2}{c}{Feature:  Avalon fingerprint} \\
\cmidrule(r){2-3} \cmidrule(r){4-5}
& Count (Ratio) & MolCLR & Count (Ratio) & MolCLR \\
\midrule
{[}0\phantom{/3}, 1/3] & \phantom{00}8 (0.0092) & 1.4442  & \phantom{00}0 (0.0000) & -  \\
{(}1/3, 2/3] & \phantom{0}34 (0.0389) & 1.1294  & \phantom{0}28 (0.0321) & 1.3299  \\
{(}2/3, \phantom{1/}1] & 831 (0.9519) & 0.6407  & 845 (0.9679) & 0.6451  \\
overall & 873 (1.0000) & 0.6671  & 873 (1.0000) & 0.6671  \\
\midrule
 & \multicolumn{2}{c}{SimilarityMeasure: Sokal similarity} & \multicolumn{2}{c}{SimilarityMeasure: Dice coefficient} \\
\midrule
{[}0\phantom{/3}, 1/3] & \phantom{0}33 (0.0378) & 1.2751  & \phantom{00}0 (0.0000) & -  \\
{(}1/3, 2/3] & 398 (0.4559) & 0.7366  & \phantom{0}33 (0.0378) & 1.2751  \\
{(}2/3, \phantom{1/}1] & 442 (0.5063) & 0.5591  & 840 (0.9622) & 0.6432  \\
overall & 873 (1.0000) & 0.6671  & 873 (1.0000) & 0.6671  \\
\midrule
 & \multicolumn{2}{c}{Aggregation: Top-3} & \multicolumn{2}{c}{Aggregation: Top-5} \\
\midrule
{[}0\phantom{/3}, 1/3] & \phantom{0}17 (0.0195) & 1.1567  & \phantom{0}24 (0.0275) & 1.3682  \\
{(}1/3, 2/3] & 171 (0.1959) & 0.8839  & 240 (0.2749) & 0.7861  \\
{(}2/3, \phantom{1/}1] & 685 (0.7847) & 0.6008  & 609 (0.6976) & 0.5926  \\
overall & 873 (1.0000) & 0.6671  & 873 (1.0000) & 0.6671  \\
\midrule
\multicolumn{5}{c}{Randomized Split (R$^2$)} \\
\midrule
\multirow{2}{*}{Bin} & \multicolumn{2}{c}{Feature: RDKit fingerprint} & \multicolumn{2}{c}{Feature:  Avalon fingerprint} \\
\cmidrule(r){2-3} \cmidrule(r){4-5}
& Count (Ratio) & MolCLR & Count (Ratio) & MolCLR \\
\midrule
{[}0\phantom{/3}, 1/3] & \phantom{00}8 (0.0092) & 0.2319  & \phantom{00}0 (0.0000) & -  \\
{(}1/3, 2/3] & \phantom{0}34 (0.0389) & 0.2253  & \phantom{0}28 (0.0321) & 0.2166  \\
{(}2/3, \phantom{1/}1] & 831 (0.9519) & 0.5899  & 845 (0.9679) & 0.5845  \\
overall & 873 (1.0000) & 0.5697  & 873 (1.0000) & 0.5697  \\
\midrule
 & \multicolumn{2}{c}{SimilarityMeasure: Sokal similarity} & \multicolumn{2}{c}{SimilarityMeasure: Dice coefficient} \\
\midrule
{[}0\phantom{/3}, 1/3] & \phantom{0}33 (0.0378) & 0.0871  & \phantom{00}0 (0.0000) & -  \\
{(}1/3, 2/3] & 398 (0.4559) & 0.5068  & \phantom{0}33 (0.0378) & 0.0871  \\
{(}2/3, \phantom{1/}1] & 442 (0.5063) & 0.6469  & 840 (0.9622) & 0.5898  \\
overall & 873 (1.0000) & 0.5697  & 873 (1.0000) & 0.5697  \\
\midrule
 & \multicolumn{2}{c}{Aggregation: Top-3} & \multicolumn{2}{c}{Aggregation: Top-5} \\
\midrule
{[}0\phantom{/3}, 1/3] & \phantom{0}17 (0.0195) & -0.0685  & \phantom{0}24 (0.0275) & -0.0237  \\
{(}1/3, 2/3] & 171 (0.1959) & 0.4390  & 240 (0.2749) & 0.4881  \\
{(}2/3, \phantom{1/}1] & 685 (0.7847) & 0.5930  & 609 (0.6976) & 0.5965  \\
overall & 873 (1.0000) & 0.5697  & 873 (1.0000) & 0.5697  \\

\bottomrule 
\end{tabular}
\end{table}

%% file: appendix_tables/random_versus_SAE_FusionDTA_ChemBERTa.tex
\begin{table}[t]
\centering
\caption{Comparison of Randomized Split and SAE (balanced) Split at IC50 for BACE1, Ki for Carbonic anhydrase I and Carbonic anhydrase II. The methods for the detailed showcase are FusionDTA and ChemBERTa.}
\begin{tabular}{c@{\hspace{2mm}}c@{\hspace{2mm}}c@{\hspace{2mm}}c@{\hspace{2mm}}c@{\hspace{2mm}}c@{\hspace{2mm}}c}
\toprule

\multicolumn{7}{c}{IC50 for Target BACE1 (MAE)} \\
\midrule
\multirow{2}{*}{Bin} & \multicolumn{3}{c}{Randomized Split}     & \multicolumn{3}{c}{SAE (balanced) Split}  \\
\cmidrule(r){2-4} \cmidrule(r){5-7}
& Count (Ratio) & FusionDTA & ChemBERTa & Count (Ratio) & FusionDTA & ChemBERTa \\
\midrule
{[}0\phantom{/3}, 1/3] & \phantom{0}10 (0.0108) & 1.3020 & 1.1440 & 309 (0.3330) & 1.2117 & 1.2503 \\
{(}1/3, 2/3] & \phantom{0}67 (0.0722) & 0.7270 & 0.6719 & 311 (0.3351) & 0.7444 & 0.6599 \\
{(}2/3, \phantom{1/}1] & 851 (0.9170) & 0.5105 & 0.5267 & 308 (0.3319) & 0.5310 & 0.5352 \\
overall & 928 (1.0000) & 0.5347 & 0.5439 & 928 (1.0000) & 0.8292 & 0.8151 \\
\midrule
\multicolumn{7}{c}{IC50 for Target BACE1 (R$^2$)} \\
\midrule
{[}0\phantom{/3}, 1/3] & \phantom{0}10 (0.0108) & 0.0422 & 0.3204 & 309 (0.3330) & -0.5113 & -0.5641 \\
{(}1/3, 2/3] & \phantom{0}67 (0.0722) & 0.5980 & 0.6325 & 311 (0.3351) & 0.4238 & 0.5787 \\
{(}2/3, \phantom{1/}1] & 851 (0.9170) & 0.6651 & 0.6446 & 308 (0.3319) & 0.7076 & 0.7235 \\
overall & 928 (1.0000) & 0.6755 & 0.6673 & 928 (1.0000) & 0.4213 & 0.4548 \\
\midrule
\multicolumn{7}{c}{Ki for Target Carbonic anhydrase I (MAE)} \\
\midrule
\multirow{2}{*}{Bin} & \multicolumn{3}{c}{Randomized Split}     & \multicolumn{3}{c}{SAE (balanced) Split}  \\
\cmidrule(r){2-4} \cmidrule(r){5-7}
& Count (Ratio) & FusionDTA & ChemBERTa & Count (Ratio) & FusionDTA & ChemBERTa \\
\midrule
{[}0\phantom{/3}, 1/3] & \phantom{00}7 (0.0079) & 0.9363 & 0.9564 & 264 (0.2983) & 1.0252 & 0.9245 \\
{(}1/3, 2/3] & 205 (0.2316) & 0.7086 & 0.7085 & 311 (0.3514) & 0.7362 & 0.7228 \\
{(}2/3, \phantom{1/}1] & 673 (0.7605) & 0.5203 & 0.5440 & 310 (0.3503) & 0.6181 & 0.6060 \\
overall & 885 (1.0000) & 0.5673 & 0.5854 & 885 (1.0000) & 0.7810 & 0.7421 \\
\midrule
\multicolumn{7}{c}{Ki for Target Carbonic anhydrase I (R$^2$)} \\
\midrule
{[}0\phantom{/3}, 1/3] & \phantom{00}7 (0.0079) & 0.0634 & 0.1421 & 264 (0.2983) & -0.4131 & -0.1076 \\
{(}1/3, 2/3] & 205 (0.2316) & 0.3536 & 0.3761 & 311 (0.3514) & 0.2106 & 0.2829 \\
{(}2/3, \phantom{1/}1] & 673 (0.7605) & 0.4500 & 0.4231 & 310 (0.3503) & 0.3532 & 0.3299 \\
overall & 885 (1.0000) & 0.4259 & 0.4161 & 885 (1.0000) & 0.1253 & 0.2334 \\
\midrule
\multicolumn{7}{c}{Ki for Target Carbonic anhydrase II (MAE)} \\
\midrule
\multirow{2}{*}{Bin} & \multicolumn{3}{c}{Randomized Split}     & \multicolumn{3}{c}{SAE (balanced) Split}  \\
\cmidrule(r){2-4} \cmidrule(r){5-7}
& Count (Ratio) & FusionDTA & ChemBERTa & Count (Ratio) & FusionDTA & ChemBERTa \\
\midrule
{[}0\phantom{/3}, 1/3] & \phantom{00}8 (0.0087) & 0.8465 & 0.5778 & 244 (0.2667) & 0.9849 & 0.9314 \\
{(}1/3, 2/3] & 201 (0.2197) & 0.6817 & 0.7419 & 342 (0.3738) & 0.8265 & 0.7390 \\
{(}2/3, \phantom{1/}1] & 706 (0.7716) & 0.5605 & 0.5997 & 329 (0.3596) & 0.6040 & 0.6072 \\
overall & 915 (1.0000) & 0.5896 & 0.6307 & 915 (1.0000) & 0.7888 & 0.7429 \\
\midrule
\multicolumn{7}{c}{Ki for Target Carbonic anhydrase II (R$^2$)} \\
\midrule
{[}0\phantom{/3}, 1/3] & \phantom{00}8 (0.0087) & 0.0349 & 0.3581 & 244 (0.2667) & 0.0488 & 0.2523 \\
{(}1/3, 2/3] & 201 (0.2197) & 0.5603 & 0.4667 & 342 (0.3738) & 0.3416 & 0.4686 \\
{(}2/3, \phantom{1/}1] & 706 (0.7716) & 0.5513 & 0.5146 & 329 (0.3596) & 0.4499 & 0.4744 \\
overall & 915 (1.0000) & 0.5570 & 0.5087 & 915 (1.0000) & 0.3583 & 0.4659 \\

\bottomrule 
\end{tabular}
\end{table}

%% file: appendix_tables/random_versus_SAE_MolCLR.tex
\begin{table}[t]
\centering
\caption{Comparison of Randomized Split and SAE (balanced) Split at IC50 for BACE1, Ki for Carbonic anhydrase I and Carbonic anhydrase II. The method for the detailed showcase is MolCLR.}
\begin{tabular}{ccccc}
\toprule

\multicolumn{5}{c}{IC50 for Target BACE1 (MAE)} \\
\midrule
\multirow{2}{*}{Bin} & \multicolumn{2}{c}{Randomized Split} & \multicolumn{2}{c}{SAE (balanced) Split}  \\
\cmidrule(r){2-3} \cmidrule(r){4-5}
& Count (Ratio) & MolCLR & Count (Ratio) & MolCLR \\
\midrule
{[}0\phantom{/3}, 1/3] & \phantom{0}10 (0.0108) & 1.2452  & 309 (0.3330) & 1.4141  \\
{(}1/3, 2/3] & \phantom{0}67 (0.0722) & 0.6952  & 311 (0.3351) & 0.6940  \\
{(}2/3, \phantom{1/}1] & 851 (0.9170) & 0.4878  & 308 (0.3319) & 0.4784  \\
overall & 928 (1.0000) & 0.5109  & 928 (1.0000) & 0.8622  \\
\midrule
\multicolumn{5}{c}{IC50 for Target BACE1 (R$^2$)} \\
\midrule
{[}0\phantom{/3}, 1/3] & \phantom{0}10 (0.0108) & -0.0492  & 309 (0.3330) & -0.9211  \\
{(}1/3, 2/3] & \phantom{0}67 (0.0722) & 0.6184  & 311 (0.3351) & 0.5107  \\
{(}2/3, \phantom{1/}1] & 851 (0.9170) & 0.6919  & 308 (0.3319) & 0.7746  \\
overall & 928 (1.0000) & 0.6974  & 928 (1.0000) & 0.3713  \\
\midrule
\multicolumn{5}{c}{Ki for Target Carbonic anhydrase I (MAE)} \\
\midrule
\multirow{2}{*}{Bin} & \multicolumn{2}{c}{Randomized Split} & \multicolumn{2}{c}{SAE (balanced) Split}  \\
\cmidrule(r){2-3} \cmidrule(r){4-5}
& Count (Ratio) & MolCLR & Count (Ratio) & MolCLR \\
\midrule
{[}0\phantom{/3}, 1/3] & \phantom{00}7 (0.0079) & 0.8510  & 264 (0.2983) & 1.0059  \\
{(}1/3, 2/3] & 205 (0.2316) & 0.6141  & 311 (0.3514) & 0.7549  \\
{(}2/3, \phantom{1/}1] & 673 (0.7605) & 0.4762  & 310 (0.3503) & 0.5755  \\
overall & 885 (1.0000) & 0.5111  & 885 (1.0000) & 0.7669  \\
\midrule
\multicolumn{5}{c}{Ki for Target Carbonic anhydrase I (R$^2$)} \\
\midrule
{[}0\phantom{/3}, 1/3] & \phantom{00}7 (0.0079) & 0.3127  & 264 (0.2983) & -0.3331  \\
{(}1/3, 2/3] & 205 (0.2316) & 0.5338  & 311 (0.3514) & 0.1585  \\
{(}2/3, \phantom{1/}1] & 673 (0.7605) & 0.5598  & 310 (0.3503) & 0.4342  \\
overall & 885 (1.0000) & 0.5572  & 885 (1.0000) & 0.1552  \\
\midrule
\multicolumn{5}{c}{Ki for Target Carbonic anhydrase II (MAE)} \\
\midrule
\multirow{2}{*}{Bin} & \multicolumn{2}{c}{Randomized Split} & \multicolumn{2}{c}{SAE (balanced) Split}  \\
\cmidrule(r){2-3} \cmidrule(r){4-5}
& Count (Ratio) & MolCLR & Count (Ratio) & MolCLR \\
\midrule
{[}0\phantom{/3}, 1/3] & \phantom{00}8 (0.0087) & 1.0278  & 244 (0.2667) & 0.8873  \\
{(}1/3, 2/3] & 201 (0.2197) & 0.6882  & 342 (0.3738) & 0.6907  \\
{(}2/3, \phantom{1/}1] & 706 (0.7716) & 0.5497  & 329 (0.3596) & 0.6232  \\
overall & 915 (1.0000) & 0.5843  & 915 (1.0000) & 0.7189  \\
\midrule
\multicolumn{5}{c}{Ki for Target Carbonic anhydrase II (R$^2$)} \\
\midrule
{[}0\phantom{/3}, 1/3] & \phantom{00}8 (0.0087) & -0.3461  & 244 (0.2667) & 0.2192  \\
{(}1/3, 2/3] & 201 (0.2197) & 0.5635  & 342 (0.3738) & 0.5347  \\
{(}2/3, \phantom{1/}1] & 706 (0.7716) & 0.5964  & 329 (0.3596) & 0.4409  \\
overall & 915 (1.0000) & 0.5856  & 915 (1.0000) & 0.4740  \\

\bottomrule 
\end{tabular}
\end{table}

%% file: iclr2025_conference.bbl
\begin{thebibliography}{71}
\providecommand{\natexlab}[1]{#1}
\providecommand{\url}[1]{\texttt{#1}}
\expandafter\ifx\csname urlstyle\endcsname\relax
  \providecommand{\doi}[1]{doi: #1}\else
  \providecommand{\doi}{doi: \begingroup \urlstyle{rm}\Url}\fi

\bibitem[Askr et~al.(2023)Askr, Elgeldawi, Aboul~Ella, Elshaier, Gomaa, and
  Hassanien]{askr2023deep}
Heba Askr, Enas Elgeldawi, Heba Aboul~Ella, Yaseen~AMM Elshaier, Mamdouh~M
  Gomaa, and Aboul~Ella Hassanien.
\newblock Deep learning in drug discovery: an integrative review and future
  challenges.
\newblock \emph{Artificial Intelligence Review}, 56\penalty0 (7):\penalty0
  5975--6037, 2023.

\bibitem[Atas~Guvenilir \& Do{\u{g}}an(2023)Atas~Guvenilir and
  Do{\u{g}}an]{atas2023approach}
Heval Atas~Guvenilir and Tunca Do{\u{g}}an.
\newblock How to approach machine learning-based prediction of
  drug/compound--target interactions.
\newblock \emph{Journal of Cheminformatics}, 15\penalty0 (1):\penalty0 16,
  2023.

\bibitem[Bajusz et~al.(2015)Bajusz, R{\'a}cz, and
  H{\'e}berger]{bajusz2015tanimoto}
D{\'a}vid Bajusz, Anita R{\'a}cz, and K{\'a}roly H{\'e}berger.
\newblock Why is tanimoto index an appropriate choice for fingerprint-based
  similarity calculations?
\newblock \emph{Journal of cheminformatics}, 7:\penalty0 1--13, 2015.

\bibitem[Bemis \& Murcko(1996)Bemis and Murcko]{bemis1996properties}
Guy~W Bemis and Mark~A Murcko.
\newblock The properties of known drugs. 1. molecular frameworks.
\newblock \emph{Journal of medicinal chemistry}, 39\penalty0 (15):\penalty0
  2887--2893, 1996.

\bibitem[Chatterjee et~al.(2023)Chatterjee, Walters, Shafi, Ahmed, Sebek, Gysi,
  Yu, Eliassi-Rad, Barab{\'a}si, and Menichetti]{chatterjee2023improving}
Ayan Chatterjee, Robin Walters, Zohair Shafi, Omair~Shafi Ahmed, Michael Sebek,
  Deisy Gysi, Rose Yu, Tina Eliassi-Rad, Albert-L{\'a}szl{\'o} Barab{\'a}si,
  and Giulia Menichetti.
\newblock Improving the generalizability of protein-ligand binding predictions
  with ai-bind.
\newblock \emph{Nature communications}, 14\penalty0 (1):\penalty0 1989, 2023.

\bibitem[Chen et~al.(2018)Chen, Engkvist, Wang, Olivecrona, and
  Blaschke]{chen2018rise}
Hongming Chen, Ola Engkvist, Yinhai Wang, Marcus Olivecrona, and Thomas
  Blaschke.
\newblock The rise of deep learning in drug discovery.
\newblock \emph{Drug discovery today}, 23\penalty0 (6):\penalty0 1241--1250,
  2018.

\bibitem[Chen et~al.(2022)Chen, Tripp, and Hern{\'a}ndez-Lobato]{chen2022meta}
Wenlin Chen, Austin Tripp, and Jos{\'e}~Miguel Hern{\'a}ndez-Lobato.
\newblock Meta-learning adaptive deep kernel gaussian processes for molecular
  property prediction.
\newblock \emph{arXiv preprint arXiv:2205.02708}, 2022.

\bibitem[Cherkasov et~al.(2014)Cherkasov, Muratov, Fourches, Varnek, Baskin,
  Cronin, Dearden, Gramatica, Martin, Todeschini, et~al.]{cherkasov2014qsar}
Artem Cherkasov, Eugene~N Muratov, Denis Fourches, Alexandre Varnek, Igor~I
  Baskin, Mark Cronin, John Dearden, Paola Gramatica, Yvonne~C Martin, Roberto
  Todeschini, et~al.
\newblock Qsar modeling: where have you been? where are you going to?
\newblock \emph{Journal of medicinal chemistry}, 57\penalty0 (12):\penalty0
  4977--5010, 2014.

\bibitem[Chithrananda et~al.(2020)Chithrananda, Grand, and
  Ramsundar]{chithrananda2020ChemBERTa}
Seyone Chithrananda, Gabriel Grand, and Bharath Ramsundar.
\newblock Chemberta: large-scale self-supervised pretraining for molecular
  property prediction.
\newblock \emph{arXiv preprint arXiv:2010.09885}, 2020.

\bibitem[Chuang et~al.(2020)Chuang, Gunsalus, and Keiser]{chuang2020learning}
Kangway~V Chuang, Laura~M Gunsalus, and Michael~J Keiser.
\newblock Learning molecular representations for medicinal chemistry:
  miniperspective.
\newblock \emph{Journal of Medicinal Chemistry}, 63\penalty0 (16):\penalty0
  8705--8722, 2020.

\bibitem[De et~al.(2022)De, Kar, Ambure, and Roy]{de2022prediction}
Priyanka De, Supratik Kar, Pravin Ambure, and Kunal Roy.
\newblock Prediction reliability of qsar models: an overview of various
  validation tools.
\newblock \emph{Archives of Toxicology}, 96\penalty0 (5):\penalty0 1279--1295,
  2022.

\bibitem[Dettmers et~al.(2024)Dettmers, Pagnoni, Holtzman, and
  Zettlemoyer]{dettmers2024qlora}
Tim Dettmers, Artidoro Pagnoni, Ari Holtzman, and Luke Zettlemoyer.
\newblock Qlora: Efficient finetuning of quantized llms.
\newblock \emph{Advances in Neural Information Processing Systems}, 36, 2024.

\bibitem[Dmitriev et~al.(2019)Dmitriev, Lagunin, Karasev, Rudik, Pogodin,
  Filimonov, and Poroikov]{dmitriev2019prediction}
Alexander~V Dmitriev, Alexey~A Lagunin, Dmitry~A Karasev, Anastasia~V Rudik,
  Pavel~V Pogodin, Dmitry~A Filimonov, and Vladimir~V Poroikov.
\newblock Prediction of drug-drug interactions related to inhibition or
  induction of drug-metabolizing enzymes.
\newblock \emph{Current Topics in Medicinal Chemistry}, 19\penalty0
  (5):\penalty0 319--336, 2019.

\bibitem[Fang et~al.(2022)Fang, Liu, Lei, He, Zhang, Zhou, Wang, Wu, and
  Wang]{fang2022geometry}
Xiaomin Fang, Lihang Liu, Jieqiong Lei, Donglong He, Shanzhuo Zhang, Jingbo
  Zhou, Fan Wang, Hua Wu, and Haifeng Wang.
\newblock Geometry-enhanced molecular representation learning for property
  prediction.
\newblock \emph{Nature Machine Intelligence}, 4\penalty0 (2):\penalty0
  127--134, 2022.

\bibitem[Gallego-Posada \& Ramirez(2022)Gallego-Posada and
  Ramirez]{gallegoPosada2022cooper}
Jose Gallego-Posada and Juan Ramirez.
\newblock {Cooper: a toolkit for Lagrangian-based constrained optimization}.
\newblock \url{https://github.com/cooper-org/cooper}, 2022.

\bibitem[Gilson et~al.(2016)Gilson, Liu, Baitaluk, Nicola, Hwang, and
  Chong]{gilson2016bindingdb}
Michael~K Gilson, Tiqing Liu, Michael Baitaluk, George Nicola, Linda Hwang, and
  Jenny Chong.
\newblock Bindingdb in 2015: a public database for medicinal chemistry,
  computational chemistry and systems pharmacology.
\newblock \emph{Nucleic acids research}, 44\penalty0 (D1):\penalty0
  D1045--D1053, 2016.

\bibitem[Guan et~al.(2023)Guan, Qian, Peng, Su, Peng, and Ma]{guan20233d}
Jiaqi Guan, Wesley~Wei Qian, Xingang Peng, Yufeng Su, Jian Peng, and Jianzhu
  Ma.
\newblock 3d equivariant diffusion for target-aware molecule generation and
  affinity prediction.
\newblock \emph{arXiv preprint arXiv:2303.03543}, 2023.

\bibitem[Harren et~al.(2024)Harren, Gutermuth, Grebner, Hessler, and
  Rarey]{harren2024modern}
Tobias Harren, Torben Gutermuth, Christoph Grebner, Gerhard Hessler, and
  Matthias Rarey.
\newblock Modern machine-learning for binding affinity estimation of
  protein--ligand complexes: Progress, opportunities, and challenges.
\newblock \emph{Wiley Interdisciplinary Reviews: Computational Molecular
  Science}, 14\penalty0 (3):\penalty0 e1716, 2024.

\bibitem[Hu et~al.(2023)Hu, Liu, Zhang, Huang, Zhang, Yu, Xiong, Liu, Ke, and
  Hong]{hu2023sam}
Zhiqiang Hu, Wenfeng Liu, Chenbin Zhang, Jiawen Huang, Shaoting Zhang, Huiqun
  Yu, Yi~Xiong, Hao Liu, Song Ke, and Liang Hong.
\newblock Sam-dta: a sequence-agnostic model for drug--target binding affinity
  prediction.
\newblock \emph{Briefings in Bioinformatics}, 24\penalty0 (1):\penalty0
  bbac533, 2023.

\bibitem[Huang et~al.(2021)Huang, Fu, Gao, Zhao, Roohani, Leskovec, Coley,
  Xiao, Sun, and Zitnik]{huang2021therapeutics}
Kexin Huang, Tianfan Fu, Wenhao Gao, Yue Zhao, Yusuf Roohani, Jure Leskovec,
  Connor~W Coley, Cao Xiao, Jimeng Sun, and Marinka Zitnik.
\newblock Therapeutics data commons: Machine learning datasets and tasks for
  drug discovery and development.
\newblock \emph{arXiv preprint arXiv:2102.09548}, 2021.

\bibitem[Hughes et~al.(2011)Hughes, Rees, Kalindjian, and
  Philpott]{hughes2011principles}
James~P Hughes, Stephen Rees, S~Barrett Kalindjian, and Karen~L Philpott.
\newblock Principles of early drug discovery.
\newblock \emph{British journal of pharmacology}, 162\penalty0 (6):\penalty0
  1239--1249, 2011.

\bibitem[Jiang et~al.(2023)Jiang, Jin, Jin, Xiao, Wu, Liu, Zhang, Zeng, Yang,
  and Niu]{jiang2023pharmacophoric}
Yinghui Jiang, Shuting Jin, Xurui Jin, Xianglu Xiao, Wenfan Wu, Xiangrong Liu,
  Qiang Zhang, Xiangxiang Zeng, Guang Yang, and Zhangming Niu.
\newblock Pharmacophoric-constrained heterogeneous graph transformer model for
  molecular property prediction.
\newblock \emph{Communications Chemistry}, 6\penalty0 (1):\penalty0 60, 2023.

\bibitem[Kalemati et~al.(2024)Kalemati, Zamani~Emani, and
  Koohi]{kalemati2024dcgan}
Mahmood Kalemati, Mojtaba Zamani~Emani, and Somayyeh Koohi.
\newblock Dcgan-dta: Predicting drug-target binding affinity with deep
  convolutional generative adversarial networks.
\newblock \emph{BMC genomics}, 25\penalty0 (1):\penalty0 411, 2024.

\bibitem[Karimi et~al.(2019)Karimi, Wu, Wang, and Shen]{karimi2019deepaffinity}
Mostafa Karimi, Di~Wu, Zhangyang Wang, and Yang Shen.
\newblock Deepaffinity: interpretable deep learning of compound--protein
  affinity through unified recurrent and convolutional neural networks.
\newblock \emph{Bioinformatics}, 35\penalty0 (18):\penalty0 3329--3338, 2019.

\bibitem[Kim et~al.(2023)Kim, Chen, Cheng, Gindulyte, He, He, Li, Shoemaker,
  Thiessen, Yu, et~al.]{kim2023pubchem}
Sunghwan Kim, Jie Chen, Tiejun Cheng, Asta Gindulyte, Jia He, Siqian He,
  Qingliang Li, Benjamin~A Shoemaker, Paul~A Thiessen, Bo~Yu, et~al.
\newblock Pubchem 2023 update.
\newblock \emph{Nucleic acids research}, 51\penalty0 (D1):\penalty0
  D1373--D1380, 2023.

\bibitem[Krishnan et~al.(2021)Krishnan, Bung, Bulusu, and
  Roy]{krishnan2021accelerating}
Sowmya~Ramaswamy Krishnan, Navneet Bung, Gopalakrishnan Bulusu, and Arijit Roy.
\newblock Accelerating de novo drug design against novel proteins using deep
  learning.
\newblock \emph{Journal of Chemical Information and Modeling}, 61\penalty0
  (2):\penalty0 621--630, 2021.

\bibitem[Krstajic et~al.(2014)Krstajic, Buturovic, Leahy, and
  Thomas]{krstajic2014cross}
Damjan Krstajic, Ljubomir~J Buturovic, David~E Leahy, and Simon Thomas.
\newblock Cross-validation pitfalls when selecting and assessing regression and
  classification models.
\newblock \emph{Journal of cheminformatics}, 6:\penalty0 1--15, 2014.

\bibitem[Landrum et~al.(2023)Landrum, Beckers, Lanini, Schneider, Stiefl, and
  Riniker]{landrum2023simpd}
Gregory~A Landrum, Maximilian Beckers, Jessica Lanini, Nadine Schneider,
  Nikolaus Stiefl, and Sereina Riniker.
\newblock Simpd: an algorithm for generating simulated time splits for
  validating machine learning approaches.
\newblock \emph{Journal of cheminformatics}, 15\penalty0 (1):\penalty0 119,
  2023.

\bibitem[Leonard \& Roy(2006)Leonard and Roy]{leonard2006selection}
J~Thomas Leonard and Kunal Roy.
\newblock On selection of training and test sets for the development of
  predictive qsar models.
\newblock \emph{QSAR \& Combinatorial Science}, 25\penalty0 (3):\penalty0
  235--251, 2006.

\bibitem[Li et~al.(2021)Li, Lu, Sze, Su, Chan, and Leung]{li2021machine}
Hongjian Li, Gang Lu, Kam-Heung Sze, Xianwei Su, Wai-Yee Chan, and Kwong-Sak
  Leung.
\newblock Machine-learning scoring functions trained on complexes dissimilar to
  the test set already outperform classical counterparts on a blind benchmark.
\newblock \emph{Briefings in bioinformatics}, 22\penalty0 (6):\penalty0
  bbab225, 2021.

\bibitem[Li et~al.(2022)Li, Wu, Wang, Li, Jiang, and Bai]{li2022recent}
Shiwei Li, Sanan Wu, Lin Wang, Fenglei Li, Hualiang Jiang, and Fang Bai.
\newblock Recent advances in predicting protein--protein interactions with the
  aid of artificial intelligence algorithms.
\newblock \emph{Current Opinion in Structural Biology}, 73:\penalty0 102344,
  2022.

\bibitem[Li \& Yang(2017)Li and Yang]{li2017structural}
Yang Li and Jianyi Yang.
\newblock Structural and sequence similarity makes a significant impact on
  machine-learning-based scoring functions for protein--ligand interactions.
\newblock \emph{Journal of chemical information and modeling}, 57\penalty0
  (4):\penalty0 1007--1012, 2017.

\bibitem[Li et~al.(2019)Li, Rezaei, Li, and Li]{li2019deepatom}
Yanjun Li, Mohammad~A Rezaei, Chenglong Li, and Xiaolin Li.
\newblock Deepatom: A framework for protein-ligand binding affinity prediction.
\newblock In \emph{2019 IEEE International Conference on Bioinformatics and
  Biomedicine (BIBM)}, pp.\  303--310. IEEE, 2019.

\bibitem[Liu et~al.(2024)Liu, Shi, Zhang, Zhang, Kawaguchi, Wang, and
  Chua]{liu2024rethinking}
Zhiyuan Liu, Yaorui Shi, An~Zhang, Enzhi Zhang, Kenji Kawaguchi, Xiang Wang,
  and Tat-Seng Chua.
\newblock Rethinking tokenizer and decoder in masked graph modeling for
  molecules.
\newblock \emph{Advances in Neural Information Processing Systems}, 36, 2024.

\bibitem[Luo et~al.(2024)Luo, Liu, Qu, Dong, and Wang]{luo2024enhancing}
Ding Luo, Dandan Liu, Xiaoyang Qu, Lina Dong, and Binju Wang.
\newblock Enhancing generalizability in protein--ligand binding affinity
  prediction with multimodal contrastive learning.
\newblock \emph{Journal of Chemical Information and Modeling}, 64\penalty0
  (6):\penalty0 1892--1906, 2024.

\bibitem[Luo et~al.(2017)Luo, Zhao, Zhou, Yang, Zhang, Kuang, Peng, Chen, and
  Zeng]{luo2017network}
Yunan Luo, Xinbin Zhao, Jingtian Zhou, Jinglin Yang, Yanqing Zhang, Wenhua
  Kuang, Jian Peng, Ligong Chen, and Jianyang Zeng.
\newblock A network integration approach for drug-target interaction prediction
  and computational drug repositioning from heterogeneous information.
\newblock \emph{Nature communications}, 8\penalty0 (1):\penalty0 573, 2017.

\bibitem[Mathai et~al.(2020)Mathai, Chen, and Kirchmair]{mathai2020validation}
Neann Mathai, Ya~Chen, and Johannes Kirchmair.
\newblock Validation strategies for target prediction methods.
\newblock \emph{Briefings in bioinformatics}, 21\penalty0 (3):\penalty0
  791--802, 2020.

\bibitem[Monteiro et~al.(2022)Monteiro, Sim{\~o}es, {\'A}vila, Abbasi,
  Oliveira, and Arrais]{monteiro2022explainable}
Nelson~RC Monteiro, Carlos~JV Sim{\~o}es, Henrique~V {\'A}vila, Maryam Abbasi,
  Jos{\'e}~L Oliveira, and Joel~P Arrais.
\newblock Explainable deep drug--target representations for binding affinity
  prediction.
\newblock \emph{BMC bioinformatics}, 23\penalty0 (1):\penalty0 237, 2022.

\bibitem[Nguyen et~al.(2021)Nguyen, Le, Quinn, Nguyen, Le, and
  Venkatesh]{nguyen2021graphdta}
Thin Nguyen, Hang Le, Thomas~P Quinn, Tri Nguyen, Thuc~Duy Le, and Svetha
  Venkatesh.
\newblock Graphdta: predicting drug--target binding affinity with graph neural
  networks.
\newblock \emph{Bioinformatics}, 37\penalty0 (8):\penalty0 1140--1147, 2021.

\bibitem[Nguyen et~al.(2022)Nguyen, Nguyen, and Tran]{nguyen2022mitigating}
Tri~Minh Nguyen, Thin Nguyen, and Truyen Tran.
\newblock Mitigating cold-start problems in drug-target affinity prediction
  with interaction knowledge transferring.
\newblock \emph{Briefings in Bioinformatics}, 23\penalty0 (4):\penalty0
  bbac269, 2022.

\bibitem[{\"O}zt{\"u}rk et~al.(2018){\"O}zt{\"u}rk, {\"O}zg{\"u}r, and
  Ozkirimli]{ozturk2018deepdta}
Hakime {\"O}zt{\"u}rk, Arzucan {\"O}zg{\"u}r, and Elif Ozkirimli.
\newblock Deepdta: deep drug--target binding affinity prediction.
\newblock \emph{Bioinformatics}, 34\penalty0 (17):\penalty0 i821--i829, 2018.

\bibitem[Pahikkala et~al.(2015)Pahikkala, Airola, Pietil{\"a}, Shakyawar,
  Szwajda, Tang, and Aittokallio]{pahikkala2015toward}
Tapio Pahikkala, Antti Airola, Sami Pietil{\"a}, Sushil Shakyawar, Agnieszka
  Szwajda, Jing Tang, and Tero Aittokallio.
\newblock Toward more realistic drug--target interaction predictions.
\newblock \emph{Briefings in bioinformatics}, 16\penalty0 (2):\penalty0
  325--337, 2015.

\bibitem[Park \& Marcotte(2012)Park and Marcotte]{park2012flaws}
Yungki Park and Edward~M Marcotte.
\newblock Flaws in evaluation schemes for pair-input computational predictions.
\newblock \emph{Nature methods}, 9\penalty0 (12):\penalty0 1134--1136, 2012.

\bibitem[Puzyn et~al.(2011)Puzyn, Mostrag-Szlichtyng, Gajewicz, Skrzy{\'n}ski,
  and Worth]{puzyn2011investigating}
Tomasz Puzyn, Aleksandra Mostrag-Szlichtyng, Agnieszka Gajewicz, Micha{\l}
  Skrzy{\'n}ski, and Andrew~P Worth.
\newblock Investigating the influence of data splitting on the predictive
  ability of qsar/qspr models.
\newblock \emph{Structural Chemistry}, 22:\penalty0 795--804, 2011.

\bibitem[Scantlebury et~al.(2023)Scantlebury, Vost, Carbery, Hadfield,
  Turnbull, Brown, Chenthamarakshan, Das, Grosjean, Von~Delft,
  et~al.]{scantlebury2023small}
Jack Scantlebury, Lucy Vost, Anna Carbery, Thomas~E Hadfield, Oliver~M
  Turnbull, Nathan Brown, Vijil Chenthamarakshan, Payel Das, Harold Grosjean,
  Frank Von~Delft, et~al.
\newblock A small step toward generalizability: training a machine learning
  scoring function for structure-based virtual screening.
\newblock \emph{Journal of Chemical Information and Modeling}, 63\penalty0
  (10):\penalty0 2960--2974, 2023.

\bibitem[Sharma \& Bhatia(2021)Sharma and Bhatia]{sharma2021recent}
Smriti Sharma and Vinayak Bhatia.
\newblock Recent trends in qsar in modelling of drug-protein and
  protein-protein interactions.
\newblock \emph{Combinatorial Chemistry \& High Throughput Screening},
  24\penalty0 (7):\penalty0 1031--1041, 2021.

\bibitem[Sheridan(2013)]{sheridan2013time}
Robert~P Sheridan.
\newblock Time-split cross-validation as a method for estimating the goodness
  of prospective prediction.
\newblock \emph{Journal of chemical information and modeling}, 53\penalty0
  (4):\penalty0 783--790, 2013.

\bibitem[Sheridan et~al.(2004)Sheridan, Feuston, Maiorov, and
  Kearsley]{sheridan2004similarity}
Robert~P Sheridan, Bradley~P Feuston, Vladimir~N Maiorov, and Simon~K Kearsley.
\newblock Similarity to molecules in the training set is a good discriminator
  for prediction accuracy in qsar.
\newblock \emph{Journal of chemical information and computer sciences},
  44\penalty0 (6):\penalty0 1912--1928, 2004.

\bibitem[Sheridan et~al.(2022)Sheridan, Culberson, Joshi, Tudor, and
  Karnachi]{sheridan2022prediction}
Robert~P Sheridan, J~Chris Culberson, Elizabeth Joshi, Matthew Tudor, and
  Prabha Karnachi.
\newblock Prediction accuracy of production admet models as a function of
  version: activity cliffs rule.
\newblock \emph{Journal of Chemical Information and Modeling}, 62\penalty0
  (14):\penalty0 3275--3280, 2022.

\bibitem[Sieg et~al.(2019)Sieg, Flachsenberg, and Rarey]{sieg2019need}
Jochen Sieg, Florian Flachsenberg, and Matthias Rarey.
\newblock In need of bias control: evaluating chemical data for machine
  learning in structure-based virtual screening.
\newblock \emph{Journal of chemical information and modeling}, 59\penalty0
  (3):\penalty0 947--961, 2019.

\bibitem[Simm et~al.(2021)Simm, Humbeck, Zalewski, Sturm, Heyndrickx, Moreau,
  Beck, and Schuffenhauer]{simm2021splitting}
Jaak Simm, Lina Humbeck, Adam Zalewski, Noe Sturm, Wouter Heyndrickx, Yves
  Moreau, Bernd Beck, and Ansgar Schuffenhauer.
\newblock Splitting chemical structure data sets for federated
  privacy-preserving machine learning.
\newblock \emph{Journal of cheminformatics}, 13:\penalty0 1--14, 2021.

\bibitem[Song et~al.(2023)Song, Chen, Wang, Chen, and Ma]{song2023double}
Yuanbing Song, Jinghua Chen, Wenju Wang, Gang Chen, and Zhichong Ma.
\newblock Double-head transformer neural network for molecular property
  prediction.
\newblock \emph{Journal of Cheminformatics}, 15\penalty0 (1):\penalty0 27,
  2023.

\bibitem[Stanley et~al.(2021)Stanley, Bronskill, Maziarz, Misztela, Lanini,
  Segler, Schneider, and Brockschmidt]{stanley2021fs}
Megan Stanley, John~F Bronskill, Krzysztof Maziarz, Hubert Misztela, Jessica
  Lanini, Marwin Segler, Nadine Schneider, and Marc Brockschmidt.
\newblock Fs-mol: A few-shot learning dataset of molecules.
\newblock In \emph{Thirty-fifth Conference on Neural Information Processing
  Systems Datasets and Benchmarks Track (Round 2)}, 2021.

\bibitem[St{\"a}rk et~al.(2022)St{\"a}rk, Ganea, Pattanaik, Barzilay, and
  Jaakkola]{stark2022equibind}
Hannes St{\"a}rk, Octavian Ganea, Lagnajit Pattanaik, Regina Barzilay, and
  Tommi Jaakkola.
\newblock Equibind: Geometric deep learning for drug binding structure
  prediction.
\newblock In \emph{International conference on machine learning}, pp.\
  20503--20521. PMLR, 2022.

\bibitem[Tang et~al.(2022)Tang, Zhong, Wang, and Zhou]{tang2022fmgnn}
Chunyan Tang, Cheng Zhong, Mian Wang, and Fengfeng Zhou.
\newblock Fmgnn: A method to predict compound-protein interaction with
  pharmacophore features and physicochemical properties of amino acids.
\newblock \emph{IEEE/ACM Transactions on Computational Biology and
  Bioinformatics}, 20\penalty0 (2):\penalty0 1030--1040, 2022.

\bibitem[Tossou et~al.(2024)Tossou, Wognum, Craig, Mary, and
  Noutahi]{tossou2024real}
Prudencio Tossou, Cas Wognum, Michael Craig, Hadrien Mary, and Emmanuel
  Noutahi.
\newblock Real-world molecular out-of-distribution: Specification and
  investigation.
\newblock \emph{Journal of Chemical Information and Modeling}, 64\penalty0
  (3):\penalty0 697--711, 2024.

\bibitem[Tricarico et~al.(2024)Tricarico, Hofmans, Lenselink, L{\'o}pez-Ramos,
  Dr{\'e}anic, and Stouten]{tricarico2024construction}
Giovanni~A Tricarico, Johan Hofmans, Eelke~B Lenselink, Miriam L{\'o}pez-Ramos,
  Marie-Pierre Dr{\'e}anic, and Pieter~FW Stouten.
\newblock Construction of balanced, chemically dissimilar training, validation
  and test sets for machine learning on molecular datasets.
\newblock 2024.

\bibitem[Tropsha et~al.(2024)Tropsha, Isayev, Varnek, Schneider, and
  Cherkasov]{tropsha2024integrating}
Alexander Tropsha, Olexandr Isayev, Alexandre Varnek, Gisbert Schneider, and
  Artem Cherkasov.
\newblock Integrating qsar modelling and deep learning in drug discovery: the
  emergence of deep qsar.
\newblock \emph{Nature Reviews Drug Discovery}, 23\penalty0 (2):\penalty0
  141--155, 2024.

\bibitem[Wan et~al.(2019)Wan, Hong, Xiao, Jiang, and Zeng]{wan2019neodti}
Fangping Wan, Lixiang Hong, An~Xiao, Tao Jiang, and Jianyang Zeng.
\newblock Neodti: neural integration of neighbor information from a
  heterogeneous network for discovering new drug--target interactions.
\newblock \emph{Bioinformatics}, 35\penalty0 (1):\penalty0 104--111, 2019.

\bibitem[Wang et~al.(2022)Wang, Wang, Cao, and
  Barati~Farimani]{wang2022molecular}
Yuyang Wang, Jianren Wang, Zhonglin Cao, and Amir Barati~Farimani.
\newblock Molecular contrastive learning of representations via graph neural
  networks.
\newblock \emph{Nature Machine Intelligence}, 4\penalty0 (3):\penalty0
  279--287, 2022.

\bibitem[Wu et~al.(2018)Wu, Ramsundar, Feinberg, Gomes, Geniesse, Pappu,
  Leswing, and Pande]{wu2018moleculenet}
Zhenqin Wu, Bharath Ramsundar, Evan~N Feinberg, Joseph Gomes, Caleb Geniesse,
  Aneesh~S Pappu, Karl Leswing, and Vijay Pande.
\newblock Moleculenet: a benchmark for molecular machine learning.
\newblock \emph{Chemical science}, 9\penalty0 (2):\penalty0 513--530, 2018.

\bibitem[Xu et~al.(2017)Xu, Wang, Zhu, and Huang]{xu2017seq2seq}
Zheng Xu, Sheng Wang, Feiyun Zhu, and Junzhou Huang.
\newblock Seq2seq fingerprint: An unsupervised deep molecular embedding for
  drug discovery.
\newblock In \emph{Proceedings of the 8th ACM international conference on
  bioinformatics, computational biology, and health informatics}, pp.\
  285--294, 2017.

\bibitem[Yang et~al.(2020)Yang, Shen, and Huang]{yang2020predicting}
Jincai Yang, Cheng Shen, and Niu Huang.
\newblock Predicting or pretending: artificial intelligence for protein-ligand
  interactions lack of sufficiently large and unbiased datasets.
\newblock \emph{Frontiers in pharmacology}, 11:\penalty0 69, 2020.

\bibitem[Yang et~al.(2019)Yang, Swanson, Jin, Coley, Eiden, Gao, Guzman-Perez,
  Hopper, Kelley, Mathea, et~al.]{yang2019analyzing}
Kevin Yang, Kyle Swanson, Wengong Jin, Connor Coley, Philipp Eiden, Hua Gao,
  Angel Guzman-Perez, Timothy Hopper, Brian Kelley, Miriam Mathea, et~al.
\newblock Analyzing learned molecular representations for property prediction.
\newblock \emph{Journal of chemical information and modeling}, 59\penalty0
  (8):\penalty0 3370--3388, 2019.

\bibitem[Yang et~al.(2022)Yang, Zhong, Zhao, and Chen]{yang2022mgraphdta}
Ziduo Yang, Weihe Zhong, Lu~Zhao, and Calvin Yu-Chian Chen.
\newblock Mgraphdta: deep multiscale graph neural network for explainable
  drug--target binding affinity prediction.
\newblock \emph{Chemical science}, 13\penalty0 (3):\penalty0 816--833, 2022.

\bibitem[Ying et~al.(2021)Ying, Leong, Alvin, and Yang]{ying2021improving}
Cheng~Zhi Ying, Chieu~Hai Leong, Liew Wen~Xing Alvin, and Chua~Jing Yang.
\newblock Improving the workflow of chemical structure elucidation with morgan
  fingerprints and the tanimoto coefficient.
\newblock In \emph{IRC-SET 2020: Proceedings of the 6th IRC Conference on
  Science, Engineering and Technology, July 2020, Singapore}, pp.\  13--24.
  Springer, 2021.

\bibitem[Yuan et~al.(2022)Yuan, Chen, and Chen]{yuan2022fusiondta}
Weining Yuan, Guanxing Chen, and Calvin Yu-Chian Chen.
\newblock Fusiondta: attention-based feature polymerizer and knowledge
  distillation for drug-target binding affinity prediction.
\newblock \emph{Briefings in Bioinformatics}, 23\penalty0 (1):\penalty0
  bbab506, 2022.

\bibitem[Zdrazil et~al.(2024)Zdrazil, Felix, Hunter, Manners, Blackshaw,
  Corbett, de~Veij, Ioannidis, Lopez, Mosquera, et~al.]{zdrazil2024chembl}
Barbara Zdrazil, Eloy Felix, Fiona Hunter, Emma~J Manners, James Blackshaw,
  Sybilla Corbett, Marleen de~Veij, Harris Ioannidis, David~Mendez Lopez,
  Juan~F Mosquera, et~al.
\newblock The chembl database in 2023: a drug discovery platform spanning
  multiple bioactivity data types and time periods.
\newblock \emph{Nucleic acids research}, 52\penalty0 (D1):\penalty0
  D1180--D1192, 2024.

\bibitem[Zhang et~al.(2022)Zhang, Wu, Yi, Zeng, Yang, Lu, Hou, and
  Cao]{zhang2022pushing}
Xiao-Chen Zhang, Cheng-Kun Wu, Jia-Cai Yi, Xiang-Xiang Zeng, Can-Qun Yang,
  Ai-Ping Lu, Ting-Jun Hou, and Dong-Sheng Cao.
\newblock Pushing the boundaries of molecular property prediction for drug
  discovery with multitask learning bert enhanced by smiles enumeration.
\newblock \emph{Research}, 2022:\penalty0 0004, 2022.

\bibitem[Zhao et~al.(2022)Zhao, Duan, Yang, Cheng, Li, and
  Wang]{zhao2022attentiondta}
Qichang Zhao, Guihua Duan, Mengyun Yang, Zhongjian Cheng, Yaohang Li, and
  Jianxin Wang.
\newblock Attentiondta: Drug--target binding affinity prediction by
  sequence-based deep learning with attention mechanism.
\newblock \emph{IEEE/ACM transactions on computational biology and
  bioinformatics}, 20\penalty0 (2):\penalty0 852--863, 2022.

\bibitem[Zhou et~al.(2023)Zhou, Gao, Ding, Zheng, Xu, Wei, Zhang, and
  Ke]{zhou2023uni}
Gengmo Zhou, Zhifeng Gao, Qiankun Ding, Hang Zheng, Hongteng Xu, Zhewei Wei,
  Linfeng Zhang, and Guolin Ke.
\newblock Uni-mol: A universal 3d molecular representation learning framework.
\newblock 2023.

\end{thebibliography}
